\pdfoutput=1

\documentclass[11pt]{article}

\usepackage{acl}

\usepackage{times}
\usepackage{latexsym}

\usepackage[T1]{fontenc}

\usepackage[utf8]{inputenc}

\usepackage{microtype}

\usepackage{inconsolata}

\usepackage{graphicx}

\usepackage{url}
\usepackage{hyperref}
\usepackage{amsmath}
\usepackage{cleveref}
\usepackage{booktabs}
\usepackage{multicol}
\usepackage{multirow}
\usepackage{graphicx}
\usepackage{array}
\usepackage{graphicx}
\usepackage{subcaption}
\usepackage{xcolor,colortbl}
\usepackage{caption}
\usepackage[compact]{titlesec}

\usepackage{xspace}  
\usepackage{pifont}  
\usepackage{enumitem}
\usepackage{amssymb} 
\newcommand{\checkbox}{\ding{111}}  
\newcommand{\radioselected}{\ding{109}}  

\newcommand{\ignore}[1]{}
\newcommand{\squishlist}{
 \begin{list}{$\bullet$}
  { \setlength{\itemsep}{0pt}
     \setlength{\parsep}{2pt}
     \setlength{\topsep}{2pt}
     \setlength{\partopsep}{0pt}
     \setlength{\leftmargin}{1em}
     \setlength{\labelwidth}{1em}
     \setlength{\labelsep}{0.4em} } }

\newcommand{\squishend}{
  \end{list}  }

          \hypersetup{
           breaklinks=true,   
           colorlinks=true,   
           pdfusetitle=true,  
        }

\newlist{checklist}{itemize}{1}
\setlist[checklist]{label=\checkbox{}}

\newlist{radiolist}{itemize}{1}
\setlist[radiolist]{label=\radioselected{}}

\newcommand{\totparticipants}{1,000}
\newcommand{\totprompts}{6,482}

\newcommand{\lengthpromptlow}{27.0}
\newcommand{\lengthpromptmiddle}{22.3}
\newcommand{\lengthpromptupper}{18.4}

\newcommand{\concretenesslow}{2.66}
\newcommand{\concretenessmiddle}{2.63}
\newcommand{\concretenessupper}{2.57}

\newcommand{\googlelow}{46.6}
\newcommand{\googlemiddle}{43.5}
\newcommand{\googleupper}{45.4}

\newcommand\blfootnote[1]{%
  \begin{NoHyper}
  \begingroup
  \hspace{-1.5em}
  \renewcommand\thefootnote{}\footnote{#1}%
  \addtocounter{footnote}{-1}%
  \endgroup
  \end{NoHyper}
}

\title{Socioeconomic Status Influences the Usage of Language Technologies}
\title{Socioeconomic Status the Interaction with Language Technologies}
\title{Socioeconomic Status Affect the Usage of Language Technologies}
\title{Socioeconomic Status Affects the Interaction with Language Technologies}

\title{Language Technologies Increase the Socioeconomic Gap}
\title{Language Technologies Widen the Socioeconomic Gap}

\title{Language Technologies are Classist}
\title{Social Class Reflects in Language Technologies Usage}

\title{Digging the Socioeconomic Gap via Language Technology Usage}
\title{Digging the Socioeconomic Gap in Language Technology Usage}
\title{AI Progress for Whom? \\ Diving the Socioeconomic Gap in Language Technology Usage}
\title{``Why don't people care about pigs''\\Socioeconomic Status Affects Individuals' Interaction with Language Technologies}
\title{``Why don't people care about pigs''\\Socioeconomic Status Affects Language Technology Interactions}
\title{The AI Gap: \\ How Socioeconomic Status Affects Language Technology Interactions}

\author{
Elisa Bassignana\textsuperscript{(*)}\\
  IT University of Copenhagen \\
  Pioneer Center for AI \\
  \texttt{elba@itu.dk} \\ \And
  Amanda Cercas Curry\textsuperscript{(*)}\\
  CENTAI Institute \\
  \texttt{amanda.cercas@centai.eu} \\\And
  Dirk Hovy\\
  Bocconi University \\
  \texttt{dirk.hovy@unibocconi.it} 
  }

\begin{document}
\maketitle
\blfootnote{\textsuperscript{(*)} Equal contribution.}

\begin{abstract}
Socioeconomic status (SES) fundamentally influences how people interact with each other and more recently, with digital technologies like Large Language Models (LLMs). While previous research has highlighted the interaction between SES and language technology, it was limited by reliance on proxy metrics and synthetic data. 
We survey \totparticipants{} individuals from \textit{diverse socioeconomic backgrounds} about their use of language technologies and generative AI, and collect \totprompts{} prompts from their previous interactions with LLMs.
We find systematic differences across SES groups in language technology usage (i.e., frequency, performed tasks), interaction styles, and topics. 
Higher SES entails a higher level of abstraction, convey requests more concisely, and topics like `inclusivity' and `travel'. 
Lower SES correlates with higher anthropomorphization of LLMs (using ``hello'' and ``thank you'') and more concrete language. 
Our findings suggest that while generative language technologies are becoming more accessible to everyone, socioeconomic linguistic differences still stratify their use to exacerbate the digital divide.
These differences underscore the importance of considering SES in developing language technologies to accommodate varying linguistic needs rooted in socioeconomic factors and limit the AI Gap across SES groups.
\end{abstract}

\section{Introduction}

\begin{figure}[t]
    \centering
    \includegraphics[width=0.9\columnwidth]{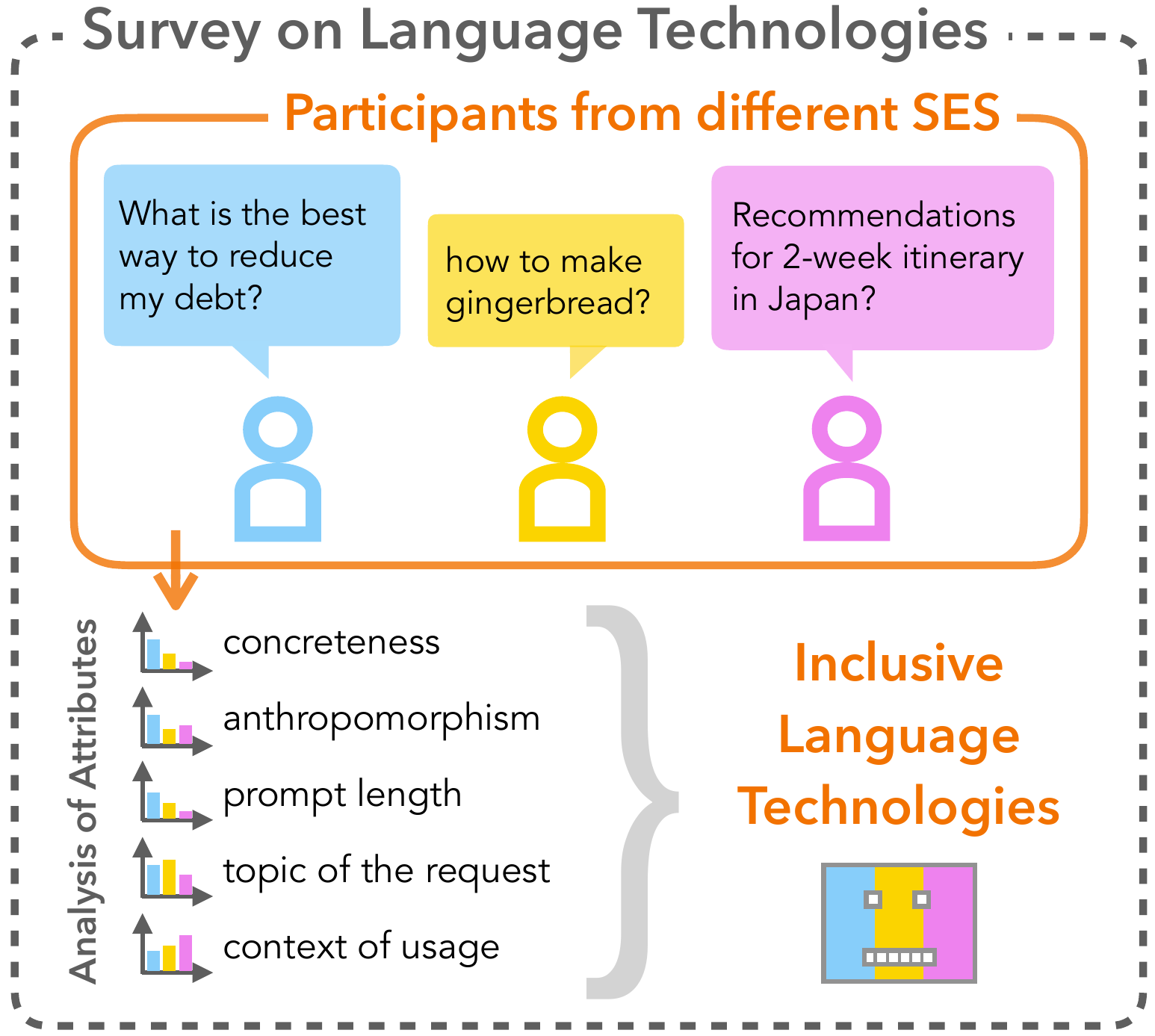}
    \caption{We survey the usage of language technologies across individual with different SES, and collect prompts from past interactions with LLMs. We find significant differences in habits and prompting strategies (see example attributes). Future language technologies should address these disparities to reduce the AI Gap. 
    }
    \label{fig:fig-1}
\end{figure}

The development of Large Language Models (LLMs), in particular ``AI chatbots'' like ChatGPT \cite{openai_chatgpt} and DeepSeek \cite{deepseekai2024deepseekv3technicalreport}, are rapidly transforming how we interact with technology. However, despite widespread accessibility, how (and how frequently) people use them varies significantly between groups. Despite their enthusiastic adoption, people from various socioeconomic backgrounds use these tools very differently. The Economist\footnote{\href{https://www.economist.com/united-states/2024/06/27/non-white-american-parents-are-embracing-ai-faster-than-white-ones}{The Economist.  Accessed 12th February 2025}} reports non-white families have adopted LLMs for education more readily than their white counterparts, and UNESCO reports women are less likely to use them at work than men \cite{unesco2023ai}. 
These differences in adoption rates (the ``AI Gap'') have raised growing concerns over their effect on exacerbating existing inequalities \cite{capraro2024impact}.
This disparity is more than just a curiosity; it is an urgent signal of a growing digital divide, similar to how educational attainment, income levels, and access to digital resources have historically driven disparities.

According to the Technology Acceptance Model (TAM; \citealp{davis1989perceived}), adoption of new technologies is influenced by their \textit{perceived usefulness} and their \textit{perceived ease of use} -- that is, users who see no practical benefits in using a technology, or who have previously had bad experiences, are less likely to adopt it. Technology adoption rates are further shaped by factors such as access, digital literacy, and ethical/privacy concerns. 
Poverty and socioeconomic status (SES) are among the main drivers of the digital divide, owing to issues of access and digital literacy \cite{mubarak2020confirming}. However, the ``digital divide'' is about more than just having access to devices and the internet; it is also about using technology effectively to improve one's life.  

As LLM use grows, a digital divide along SES lines could exacerbate inequalities in several ways.
LLMs trained on data from higher SES usage patterns may be less effective or biased against lower SES language styles and content interests, perpetuating social biases. As a result, lower SES perspectives may be underrepresented or misrepresented, resulting in skewed narratives that do not accurately reflect society.
People who use simpler language and more concrete terms may receive less effective results if models perform better with abstract or sophisticated prompts. The assistance quality could suffer, and this may not be reflected in evaluation benchmarks \cite{liao2023rethinking}.
Reduced user satisfaction could further discourage adoption among lower SES groups, widening the AI Gap.

Previous work has shown that the \textit{performance} of NLP systems is affected by sociodemographic linguistic differences, including native language \cite{reusens2024native}, race \cite{blodgett2024llms}, and social class \cite{curry-etal-2024-classist}. 
However, none of those works have examined how SES influences the use cases of language technologies or whether there are any systematic linguistic differences in the interactions with LLMs. 
Our research addresses this gap by investigating how different SES groups take advantage of current language technologies and how they interact with LLMs differently.
To the best of our knowledge, we are the first to compare the use of language technologies in general and LLMs in particular across socioeconomic groups. We discovered statistically significant differences across SES groups regarding language technology adoption and specific uses.

\paragraph{Contributions.}
Our contributions are:
\squishlist
    \item A survey of language technology use across sociodemographic groups;
    \item The first dataset of real prompts annotated with fine-grained sociodemographic information, including SES;\footnote{\url{https://huggingface.co/datasets/MilaNLProc/survey-language-technologies}}
    \item A quantitative and qualitative analysis of the differences between SES groups with respect to the use of language technologies.
\squishend
Our findings expand current research on the AI Gap, and on understanding LLM usage among the general public, contributing to the development of more inclusive language technologies.

\section{Related Work}
Interest in AI and public attitudes towards it has skyrocketed in recent years, prompting several surveys \cite[e.g.,][]{scantamburlo2024artificial} investigating how people use and perceive AI and LLMs. 
Digital divides across sociodemographic groups have been areas of concern for some time now, inspiring several studies about the desiderata of different groups \cite[e.g.,][]{blaschke-etal-2024-dialect,lent-etal-2022-creole}. 
\newcite{curry-etal-2024-classist} show that NLP systems' performance may be affected by the socioeconomic status of speakers using film and TV shows. 
\newcite{daepp2024emerging} analyze the most common intents towards chatGPT in different regions of the US and find evidence of an emerging AI Gap between regions with higher and lower incomes. 
Recently, several large-scale collections of prompts and interactions have been collected to understand the broad applications of LLMs, such as \citet{kirk2024prism}, \newcite{trippas2024users}, \newcite{zheng2023lmsyschat1m},  Huggingface's ShareGPT datasets,\footnote{\href{https://huggingface.co/collections/bunnycore/sharegpt-datasets-66fa831dcee14c587f1e6d1c}{ShareGPT Datasets}} and \newcite{wildchat}. However, none of these have have collected information on SES.
We fill the gap by surveying the usage of language technologies in general and of LLMs, collect prompts from previous interaction of the participants with AI chatbots, and analyze differences across SES groups. We highlight the importance of exploring differences across SES groups to limit the AI Gap.

\section{Socioeconomic Status}
\label{sec:social-class}

Socioeconomic status (SES) refers to an individual or group's social and economic position. A person's SES\footnote{There are different social stratification systems depending on the culture of reference (e.g., the Indian caste system, Indigenous American clans, or tribes). Our study focuses on U.S. and U.K. speakers, so we refer to the Western European class model. For a broader discussion on the intricacies of social stratification systems, see \cite{savage2016social}.} is a function of their economic, social, and cultural capital -- factors such as income, education, occupation, and wealth typically influence the SES of an individual. Still, they are often insufficient to determine it \cite{bourdieu1987makes}. SES influences almost all aspects of an individual's life: hobbies, social circle, access to experiences, and even language~\cite{Labov_2006,savage2016social}.
People’s perception of where they stand regarding socioeconomic status has important psychological effects, supporting the idea that subjective class is an important measure of socioeconomic status~\cite{cercas-curry-etal-2024-impoverished}.
To assess SES, the typical setup is to ask participants to place themselves on the socioeconomic ladder ranging from one to ten following the Macarthur scale~\cite{Adler2000RelationshipOS} where higher levels represent those who are more privileged. 

\paragraph{SES and the Digital Divide:}
SES impacts the digital divide by shaping access to technology, digital skills, and the ability to leverage digital tools for economic and social mobility \cite{mubarak2020confirming}. Higher SES affords access to better devices and paid AI services. 
Cultural capital and \textit{habitus} influence how people engage with technology -- higher social classes may develop advanced digital literacy, using AI for learning and professional growth.  
These disparities may reinforce existing inequalities, as those with greater digital access and skills gain further advantages in education, employment, and social influence. With this respect, \citet{capraro2024impact} posit that generative AI will widen the already existing digital divide.

\section{Survey Setup}
\label{sec:survey}

In our survey, we include three types of question: Sociodemographic information, inquiries about the usage of language technologies, and prompt collections (see Figure \ref{fig:fig-1}).\footnote{The whole survey can be found in \Cref{app:survey}.}
The first section includes 17 questions that aim to collect basic demographics (such as age and gender) as well as information about the socioeconomic background. We ask about participants' perception of their SES with respect to the Macarthur scale, as well as other individual factors such as level of education, parents' occupation and hobbies. Subjects could opt out of supplying this information. All information was treated in compliance with GDPR, in that subjects are fully anonymized (we have no way of connecting information to subjects, and the combination of features could not identify individuals). See also Section Ethical Considerations.
While we cannot individually verify the sociodemographic information, we can match the information provided in our survey with the demographic profile they provide in Prolific. We find that less than 2.5\% of the participants provide conflicting information (regarding gender, ethnicity or age).

For the second part of the survey, we are inspired by~\citet{lent-etal-2022-creole} and give a broad definition of ``language technologies'' before asking participants about their experience these technologies (e.g., Which of the following language technologies have you used?).\footnote{\textit{Language technology refers to any piece of software that is intended to assist humans with language specific tasks in a technological setting (i.e., on a mobile phone, tablet, computer, the internet, smart devices). Some examples of language technologies include: Spell checkers in e-mail helps people to write more professional e-mails; Google Translate helps people to translate text from one language to another; internet search engines (e.g. Google, Bing, Yahoo) help people to find websites relevant to a given query.} \cite{lent-etal-2022-creole}} Additionally, the second part includes questions which are more specific to the usage of LLMs, defined as ``AI chatbots like ChatGPT or other similar chatbots''. We ask about the frequency of usage, the applications (e.g., coding, brainstorming, writing) and the contexts of usage (e.g., working, learning, personal).
For all the questions in the first and second sections, participants could select ``Other'' or ``Prefer not to say''.

In the last section, we focus on the LLMs' usage and ask participants to provide the last ten prompts used in their interaction with AI chatbots (participants are free to look at their chat log).

\subsection{Pilot Studies}
To refine our survey, we conduct three pilot studies with 20, 20, and 79 participants, respectively. These pilots served two main purposes: first, to test the technical robustness of our self-implemented \texttt{streamlit} app\footnote{\url{https://streamlit.io/}} hosting the survey, and second, to evaluate the clarity and constraints of our questions. Based on insights from the pilots, we refined some aspects of the survey. For instance, we introduce a requirement for participants who report using AI chatbots ``every day'' or ``nearly every day'' to provide at least five prompts. Additionally, we adjusted the wording of ``please provide us with the last ten \textit{prompts} you used for your chosen AI chatbot'' to ``please provide us with the last ten \textit{questions or requests} you used for your chosen AI chatbot'' to improve clarity. After the third pilot ran smoothly, we proceeded with the large-scale study.

\subsection{Coverage}
We distributed our survey using Prolific, a crowdsourcing platform with a wide and diverse participant pool from all around the globe.\footnote{\url{https://www.prolific.com/}} The platform allows the selection of participants based on an extensive range of fine-grained demographic criteria. Crowdsourcing platforms are limited in terms of the population reached, but we still observe significant differences within the socioeconomic spectrum of individuals on Prolific (see \Cref{sec:usage-language-technologies,sec:analysis-prompts}). We expect these differences to be more pronounced in a real-world population distribution.

We selected participants to be English native speakers.
Except for one of our pilot studies, where we only constrained participants based on their first language, all other pilots and the large-scale study additionally required participants to reside in the United Kingdom (UK) or the United States (US).
While we acknowledge that opening the participation to more countries and languages would give a broader perspective on the usage of language technologies, we decide to restrict our study to participants located in the UK and in the US. Linguistic differences between groups are not uniform across all English-speaking populations, nor are available models and resources. The complexity and the cultural dependency that would need to be addressed in a wider setup. Given our available resources, our sample would have not being representative. 

We conducted the large-scale study in two phases. The first phase included 501 participants, while the second phase focused on a targeted sample of 380 participants from the low and upper social strata (see definition in \Cref{sec:social-class}). This targeted approach was necessary due to the high representation of middle-class individuals in the initial round.

\section{Results}
\label{sec:demographics}

\begin{figure}
    \centering
    \includegraphics[width=\columnwidth]{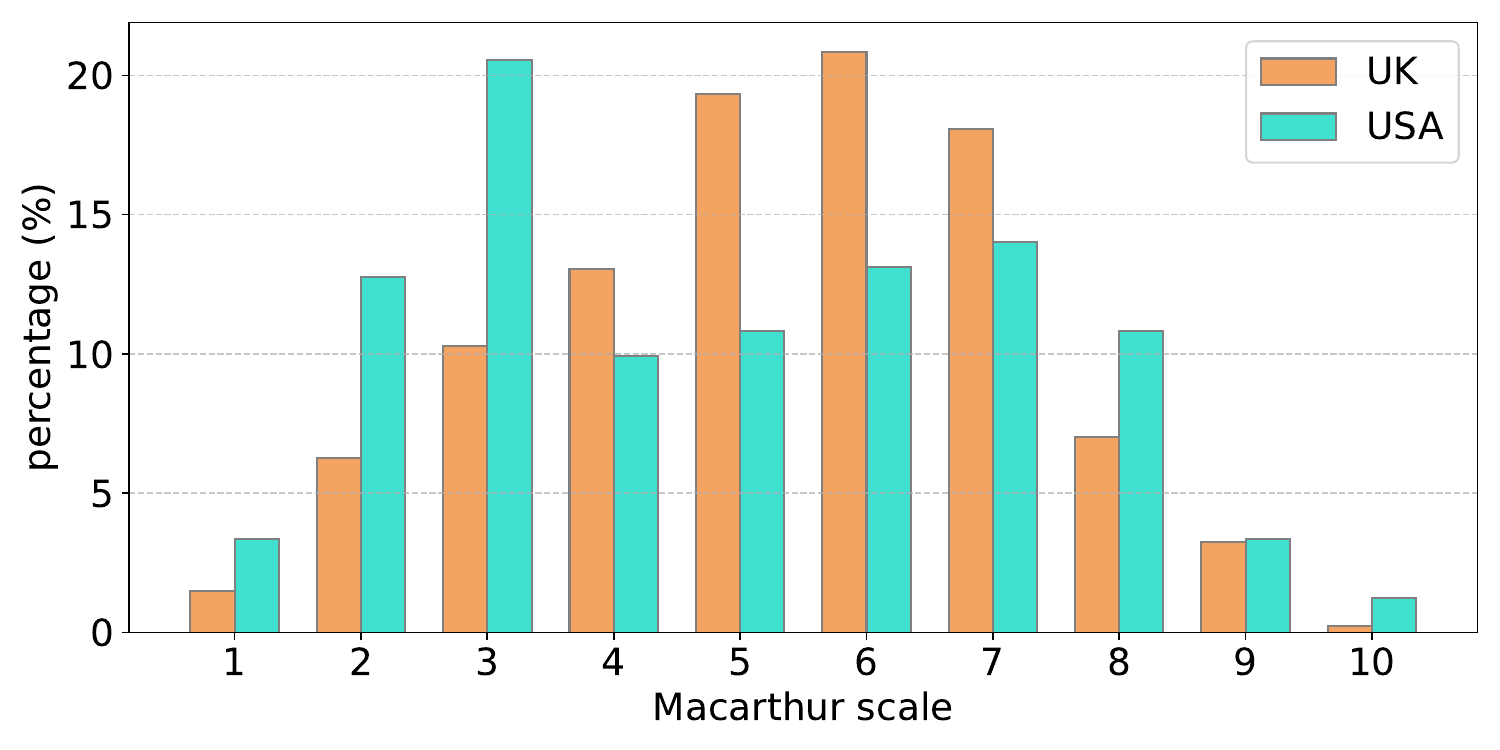}
    \caption{Distribution of participants according to the Macarthur scale~\cite{Adler2000RelationshipOS}.}
    \label{fig:distribution-ses}
\end{figure}

We collect a total of \totparticipants{} answers to our survey. 
In \Cref{fig:distribution-ses} we report the distribution of our participants over the Macarthur scale.
For this study we refer to the self-reported SES and map the Macarthur scale's values to the Western hierarchical class system (see \Cref{sec:social-class}) by binning 1-3 into lower, 4-7 middle, and 8-10 into upper. We also collect data about other socioeconomic factors that have been previously used in NLP research as proxy information for estimating SES, such as education \cite{cercas-curry-etal-2024-impoverished}. We hope this will encourage future research in NLP to consider socioeconomic status, either self-reported or via proxy factors, depending on the use case. We report in \Cref{app:demographics} the statistics about the level of education of the participant and of their parents, their current employment, the occupation of the participant and of the parents according to the European Skills, Competences, Qualifications and Occupations taxonomy (ESCO;~\citealp{esco}), the housing status (i.e., whether they own or rent) and their hobbies (following the list provided by Great British Class Survey;~\citealp{gbcs}).
Additionally, in \Cref{app:demographics} we report more detailed demographics of our participants including their gender, age, nationality, the ethnicity, the marital status and religion.

\subsection{Usage of Language Technologies}
\label{sec:usage-language-technologies}

\begin{figure}
    \centering
    \includegraphics[width=\columnwidth]{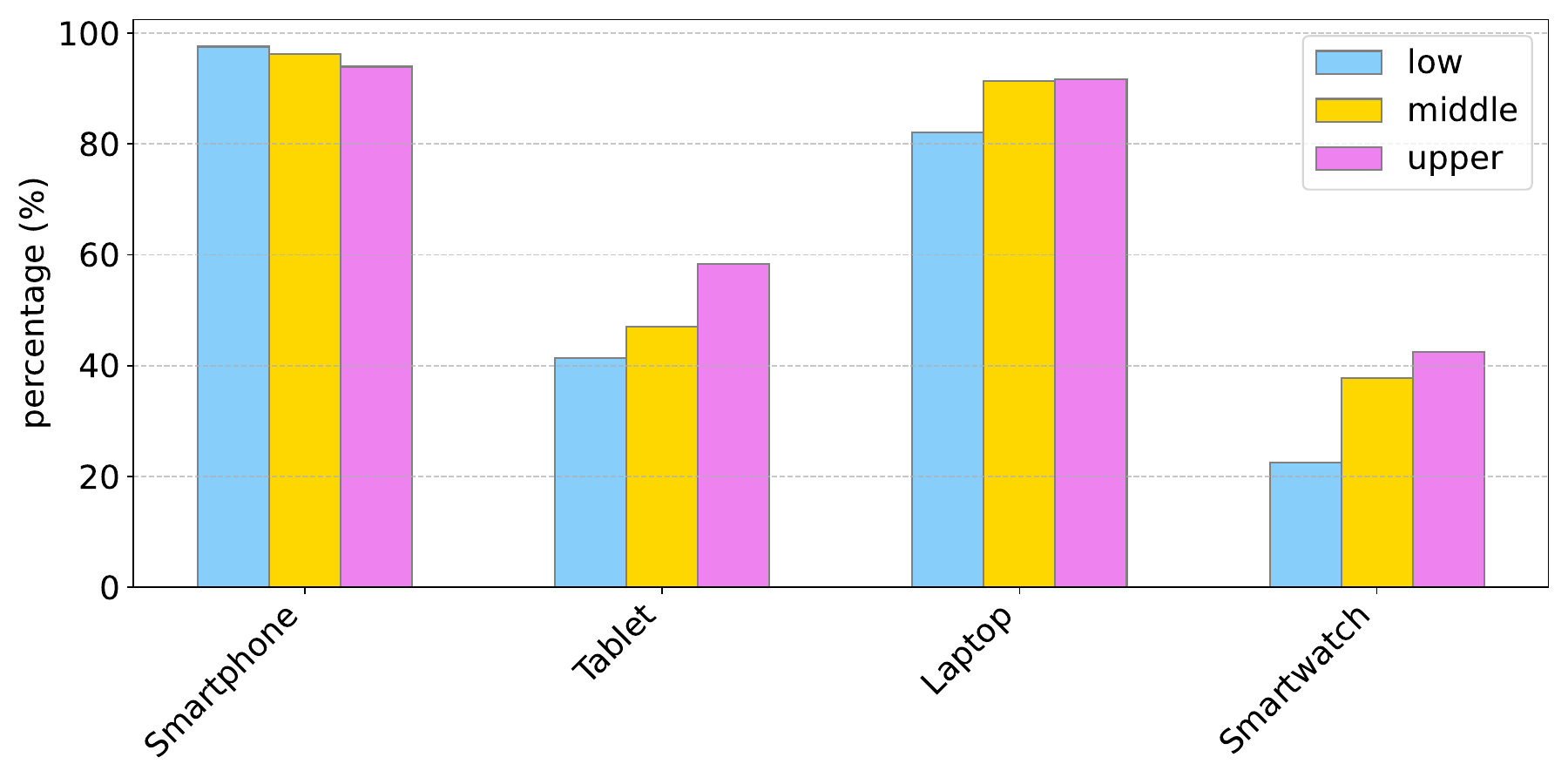}
    \caption{Daily access to digital devices from low, middle, upper classes.}
    \label{fig:tech_by_ses}
\end{figure}

First, we analyze the daily access to digital devices (i.e., smartphone, tablet, laptop, smartwatch) by individuals from low, middle and upper social classes, to check if there are significant differences in the way people may access AI chatbots. \Cref{fig:tech_by_ses} shows that the percentage of daily access to smartphones is similar for all social classes. Differences appear instead in the daily access to tablets, laptops and smartwatches, with an usage increase for higher classes (middle and upper). The differences in daily access to digital devices are statistically significant across the three social classes, as indicated by a chi-square test of independence, $\chi^2$ (df = 8, N = 2739) = 55.11, \textit{p} < 0.001.\footnote{Note that this is a multiple-choice question.} This suggests a strong association between socioeconomic status and daily access of digital devices.

\begin{figure}
    \centering
    \includegraphics[width=\columnwidth]{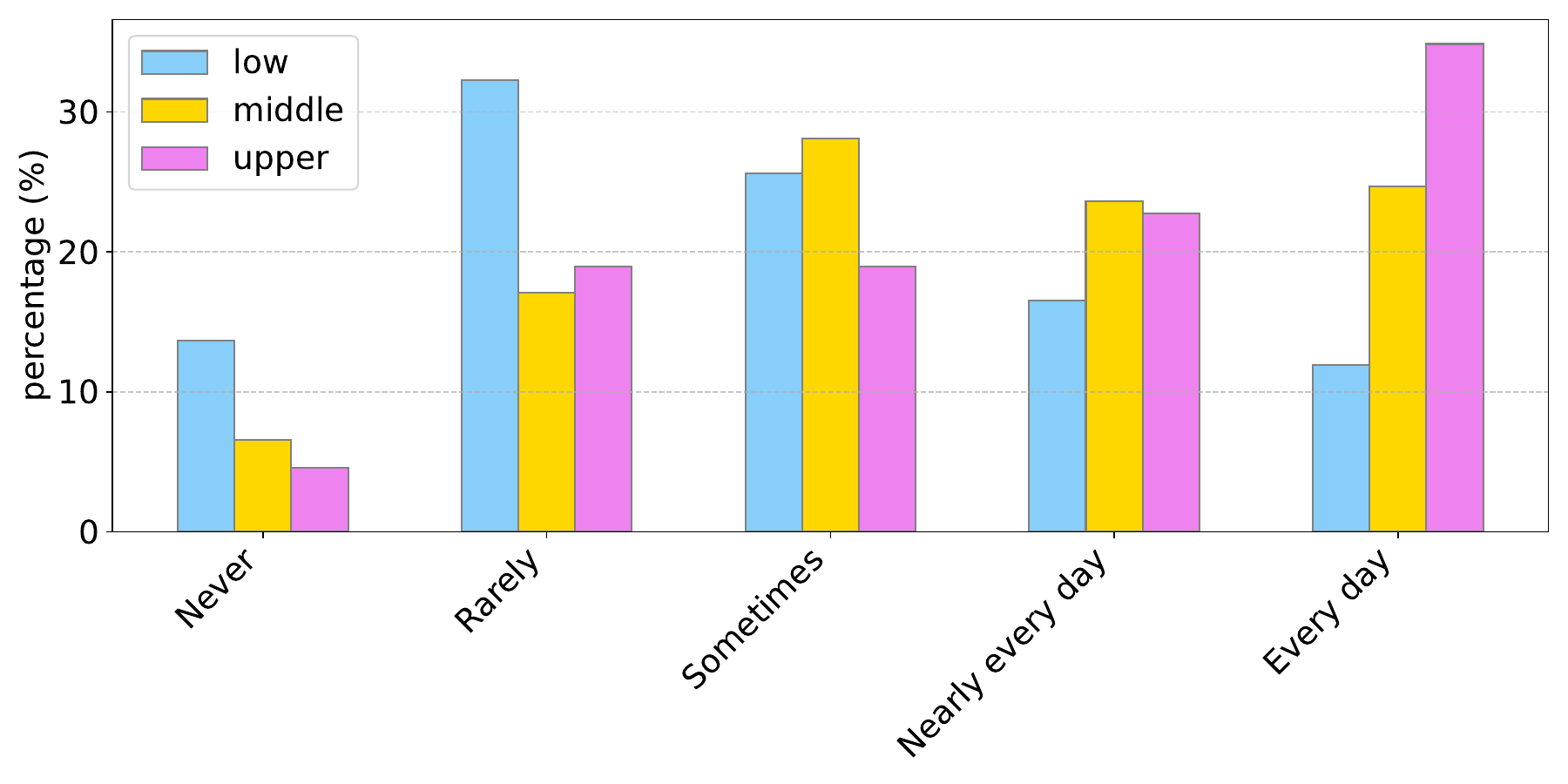}
    \caption{Frequency of usage of AI chatbots by individual from low, middle, upper classes.}
    \label{fig:frequency_by_ses}
\end{figure}

In \Cref{app:language-technologies} we report the broad statistics relative to the type of language technologies mostly used by individuals from different SES (e.g., spell checkers, dialogue systems, speech-to-text etc.). 
Below instead we focus on the usage of AI generative chatbots, like ChatGPT.
We identify a reverse trend in the frequency of usage of AI chatbots across people from low and upper social classes (see \Cref{fig:frequency_by_ses}). Individuals from low socioeconomic backgrounds (values from 1 to 3 on the Macarthur scale) tend to use the chatbots less often--see decreasing values from `Rarely' to `Every day' in \Cref{fig:frequency_by_ses}. On the other side, individuals which placed themselves from 8 to 10 on the Macarthur scale have increasing values from `Never' to `Every day'. These differences are statistically significant, as indicated by a chi-square test of independence, $\chi^2$ (df = 8, N = 997) = 67.79, \textit{p} < 0.001, suggesting a strong association between SES and frequency of usage of AI chatbots.

\begin{figure*}
    \centering
    \includegraphics[width=\textwidth]{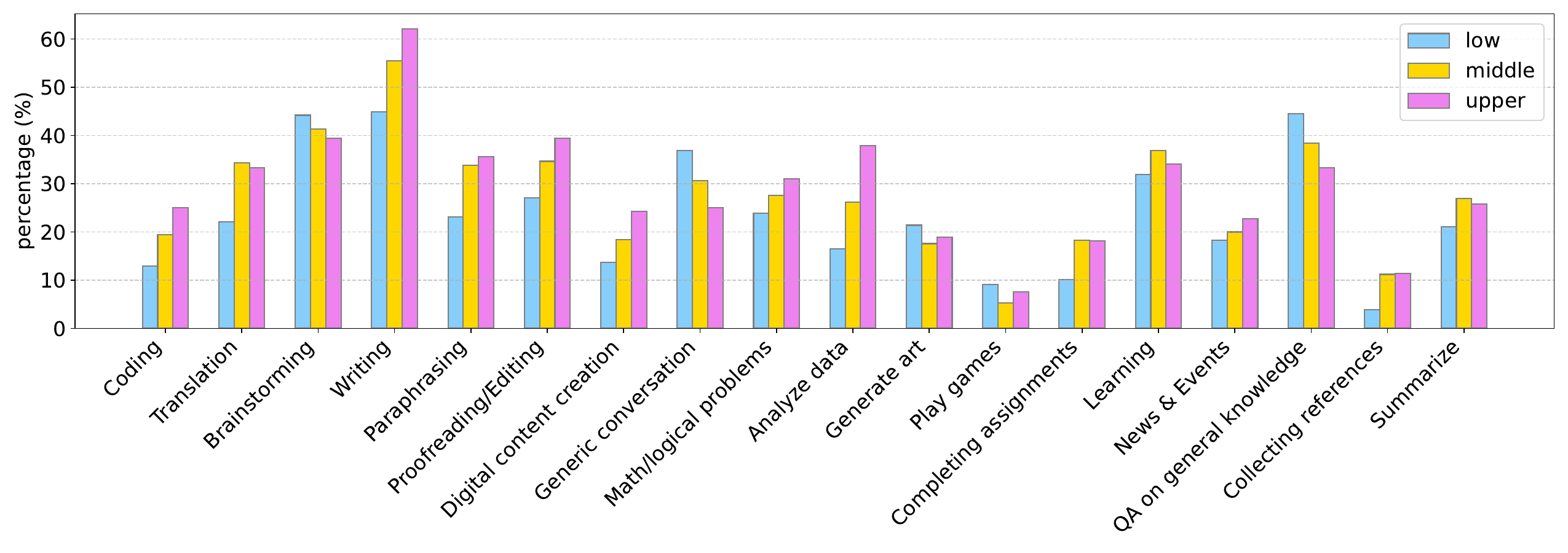}
    \caption{Tasks performed by individual from low, middle, upper classes with AI chatbots.}
    \label{fig:usecases_by_ses}
\end{figure*}

\begin{figure}
    \centering
    \includegraphics[width=\columnwidth]{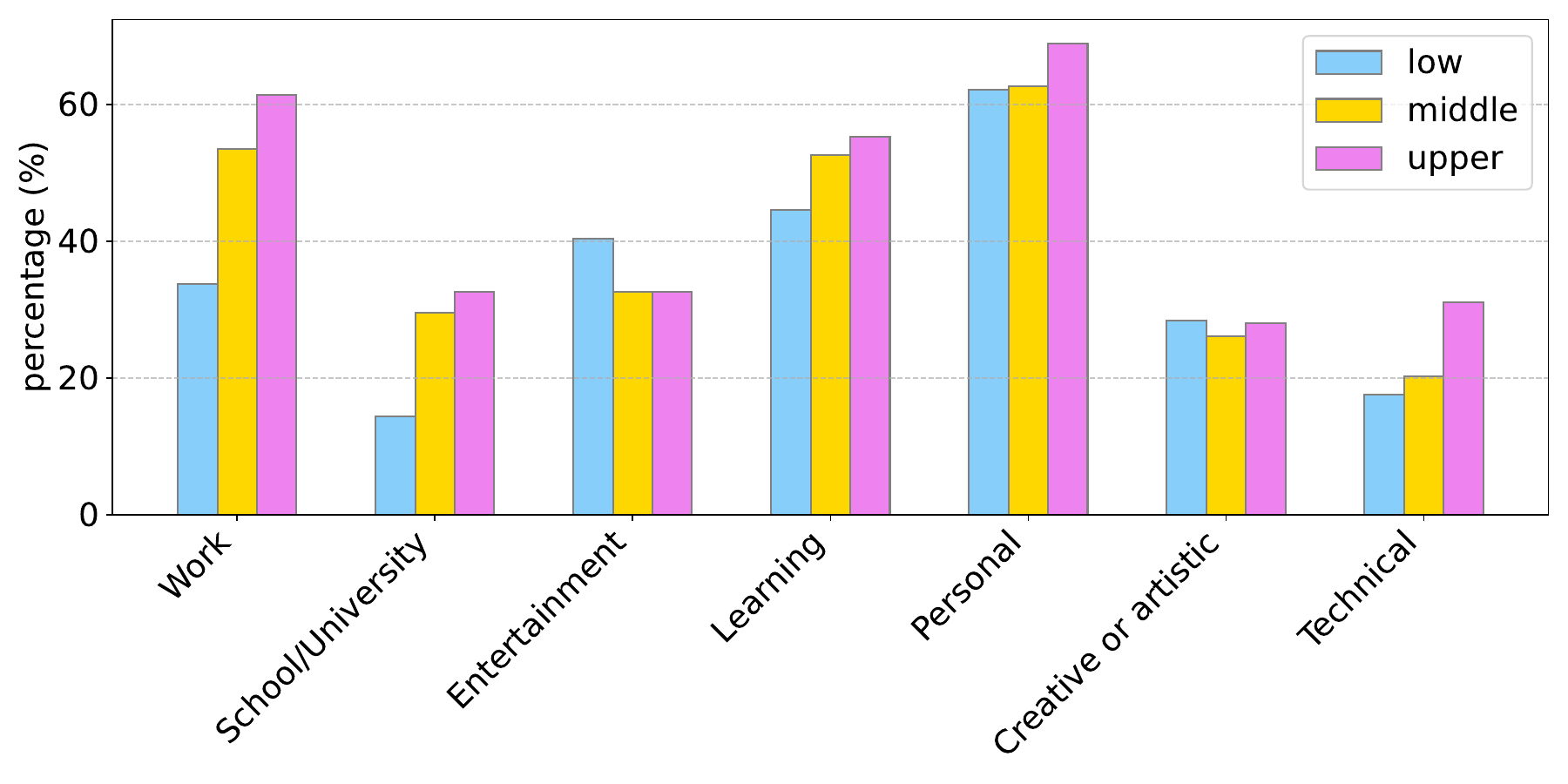}
    \caption{Context of usage of AI chatbots by individual from low, middle, upper classes.}
    \label{fig:context_by_ses}
\end{figure}

In our analysis about the context of usage of AI chatbots (see \Cref{fig:context_by_ses}) we identify a distinct higher usage in the contexts of `Work', `School/University' and `Learning' by individuals from the middle and upper classes. 
Additionally, the upper class uses AI chatbots more often in the `Personal' and in the `Technical' context, where a qualitative analysis revealed a consistent chunck of questions about machine learning. 
On the other side, individuals from the low class tend to interact more with AI chatbots in the contexts of `Entertainment'. These differences are statistically significant, as indicated by a chi-square test of independence, $\chi^2$ (df = 14, N = 2717) = 46.65, \textit{p} < 0.001,\footnote{Note that this is a multiple-choice question.} suggesting a significant association between SES and the context of AI chatbot usage.

Last, we ask more specifically about the tasks performed with the AI chatbots, and report the statistics in \Cref{fig:usecases_by_ses}. We identify that people from higher social classes (i.e., middle and upper) use the LLMs more frequently for writing and writing related tasks (i.e., paraphrasing, proofreading/editing, summarizing), and for more technical tasks (i.e., coding, solving mathematical and logical problems, analyzing data). On the other hand, individuals from the low class tend to use the AI chatbots for more generic tasks like brainstorming, generic chatbot conversations and answering questions about general knowledge. We perform a chi-squared statistical test, $\chi^2$ (df = 36, N = 4809) = 88.40, \textit{p} < 0.001\footnote{Note that this is a multiple-choice question.} revealing a strong association between SES and tasks performed with AI chatbots.

\subsection{Linguistic Analysis of the Prompts}
\label{sec:analysis-prompts}

We collect a total of \totprompts{} real prompts used by our participants in previous interactions with LLMs. We perform a linguistic analysis to investigate key characteristics, with a focus on how prompting strategies differ among individuals with varying socioeconomic backgrounds. 

\paragraph{Prompt length.}
We find differences in the average length of prompts written by people from different SES. Specifically, individuals from higher social classes tend to write shorter and more concise prompts. The average length in terms of number of words is \lengthpromptlow{} for low, \lengthpromptmiddle{} for middle and \lengthpromptupper{} for upper class. 
A bootstrap resampling significance testing indicates a statistically significant difference between the low and upper classes (\textit{p} < 0.05).\footnote{We use the implementation by~\citet{ulmer2022deep}.}
We speculate this divergence to be a consequence of higher class individuals having a wider vocabulary, which allows them to express themselves using less words.\footnote{We test several metrics for computing the vocabulary diversity across social strata (including TTR and entropy), but do not find statistically significant differences. We attribute this to the restricted nature of the data itself, which consists of prompts, that inherently limit the range of vocabulary used.}

\paragraph{Concreteness.}
\citet{bernstein-language} posits that people from higher class families are more encouraged to use language for abstract thinking in contrast to people from lower class families, which are ``limited'' to more concrete concepts.
To asses the level of concreteness and abstraction of the collected prompts, we use the list of 40,000 common English words proposed by \citet{Brysbaert2014ConcretenessRF}. In this collection, each word is evaluated  on a scale from one (abstract) to five (concrete) by at least 25 participants. The average concreteness score of the prompts written by people from the low social class is \concretenesslow{}, while it decreases (the lower the more abstract) to \concretenessmiddle{} and \concretenessupper{} respectively for people from the middle and upper class. 
While these differences are relatively small, the upper class differs significantly from the others based on bootstrap resampling (\textit{p} < 0.05).
\citet{bernstein-language}'s theory on the different level of concreteness of the language used by people from different socioeconomic background is reflected in the way people interact with LLMs: People from the upper social class interact with a more abstract language.

\paragraph{Logistic Regression Classifier.}

From the analysis above, we suspect that the prompts written by individual from different socioeconomic background are significantly different and easily distinguishable. 
To confirm our hypothesis, we train a simplistic bag-of-word classifier, which achieves a Macro-F1 score of 39.25, compared to a majority baseline of 25.02 Macro-F1.
Since we split our data into train and evaluation sets using random seeds, we apply the Almost Stochastic Order test \citep{del2018optimal, dror-etal-2019-deep} as 
implemented by \citet{ulmer2022deep} with a confidence level $\alpha = 0.05$ to assess that the simplistic bag-of-word classifier is stochastically dominant over the majority baseline ($\epsilon_\text{min} = 0$).

\begin{table*}[t]
\centering
\renewcommand{\arraystretch}{1.2} 
\resizebox{\textwidth}{!}{
\small
    \begin{tabular}{c|p{0.30\textwidth}|p{0.30\textwidth}|p{0.30\textwidth}}
    \toprule
    & \textbf{Low} & \textbf{Middle} & \textbf{Upper} \\
    \midrule
    \multirow{6}{*}{\rotatebox{90}{Finance}} 
    
    & \cellcolor{black!10} What is the cheapest place to live in the US? 
    & \cellcolor{black!10} write a business plan for an agrucultural based business 
    & \cellcolor{black!10} What are the current crypto market trend \\
    
    &  Do I need to be a member of a credit union to apply for their loans 
    & here are my monthly earnings and spends , show me how to save money 
    & when should I buy a house \\
    
    & \cellcolor{black!10} What is the best way to reduce my debt 
    & \cellcolor{black!10} How do I start a successful small business? 
    & \cellcolor{black!10} how long should I invest for?\\
    \midrule
    
    \multirow{6}{*}{\rotatebox{90}{Job}} 
    
    & What are some WFH jobs that require no experience or degrees.
    & Please can you write an application letter to a school for work experience. 
    & Create a cover letter for a new role as a communication manager \\
    
    & \cellcolor{black!10} Please write an email to get some experience and/or paid employment 
    & \cellcolor{black!10} Please write a covering letter to an employer which summarises my skills. 
    & \cellcolor{black!10} Can you suggest some effective leadership strategies for managing a team \\
    
    & Write me a covering letter for a restauraunt job 
    & Write a cover letter for assistant accountatn to match the below job description. 
    & How can I improve my email writing to sound more professional?\\
    
    \midrule
    \multirow{7}{*}{\rotatebox{90}{Food}} 
    & \cellcolor{black!10} Hello...what is the best way for me to reheat my Buffalo Wild Wings? 
    & \cellcolor{black!10} What are some creative uses for leftover rice? 
    & \cellcolor{black!10} Give me some healthy meal ideas I could cook for a family of three.  \\
    
    & give me new recipes 
    & what can I cook with leftover pork 
    & What wine with red tuna belly \\
    
    & \cellcolor{black!10} I don’t have peas and carrots 
    & \cellcolor{black!10} how to make gingerbread 
    & \cellcolor{black!10}  What's the best filtered water pitcher? Preferably one that eliminates any of the following: PFAS, PFOAS, microplastics and/or nanoplastics \\
    \bottomrule
\end{tabular}}
\caption{Example of prompts written by individuals from different social classes, divided by macro-topics.}
\label{tab:clustering}
\end{table*}

\subsection{Clustering of the Prompts}
\label{sec:clustering}

\paragraph{Methodology.}
To analyze the topics addressed in the collection of prompts, we perform topic modeling. Specifically, we (1) embed the prompts using SentenceTransformer~\cite{reimers-gurevych-2019-sentence}\footnote{We use \texttt{all-MiniLM-L6-v2}.} and M3-Embedding~\cite{chen-etal-2024-m3},\footnote{We use \texttt{bge-large-en-v1.5}} (2) cluster them using UMAP~\cite{umap} and HDBSCAN~\cite{mcinnes2017hdbscan}, (3) assign them a short and distinct description using GPT-4~\cite{openai2024gpt4technicalreport}\footnote{We use GPT-4o.} and (4) manually evaluate the clusters and the descriptions.

\paragraph{Clustering results.}
We find commonalities in the topics identified across the three social classes. These include for example \textit{translation} (e.g., ``Translate Good morning into Japanese.'', ``translate to dutch `I am fine thank you'''), \textit{mental health} (e.g., ``How do i best overcome social anxiety ?'', ``Give me advice on living with depression''), \textit{medical advice} (e.g., ``what are the dangers of etopic pregnancy'', ``Why does my 1 year old have a running nose all the time''), \textit{writing} (e.g., ``write an essay of about the dangers of using AI to generate text'', ``Hi I'm in need of some ideas for what to write in my wife's birthday card.'') and \textit{text editing} (e.g., ``Rewrite this text in formal English.'', ``Please rephrase the below email to sound more professional and authoritive'').

We also find some topics to be distinctive for specific social classes. This is especially valid for the upper class, where we identify a cluster related to \textit{travel destinations} (e.g., ``I am planning to travel for a vacation in Japan, do you have any recommendations?'', ``4 week itineray for seniors, travelling in Vietnam'') and several clusters related to abstract concepts like \textit{inclusivity} (e.g., ``How can we create a more inclusive environment for all genders?'', ``What are some specific actions, practices, or class features that make you feel most supported and valued as an LGBTQ+ student in a college course environment?'') and \textit{good communication} (e.g., ``How can we improve internal communication across different departments within our organization?'', ``Barriers of good communication''). 

Finally, some topics are addressed by individual from the whole socioeconomic spectrum, but within varying framings (see examples in \Cref{tab:clustering}). Among these, the most prominent is \textit{finance}. While within the low class we find advices for money saving, in the corresponding upper class cluster there are requests for investments advice. We also identify a common \textit{job} cluster across all three social classes. Within this, the requests from the middle class mostly involve job applications for specific positions, the ones from the low class suggestions for more generic low-skilled jobs, and the ones from the upper class often imply an high-level job positions of the user. Last, we identify a cluster of prompts related to \textit{food}. Here, the upper class is the most distinct with requests specifically targeting healthy and expensive dishes.

\subsection{User Perceptions}
User perceptions of a system, whether they perceive it as a tool or something else, affect how they interact with and use it \cite{delcker2024first}. Cues suggesting humanlikeness (such as the use of natural language) trigger social scripts and a mental model of a system with humanlike qualities \cite{reeves1996media}. At a meta-level, human metaphors (such as deep \textit{learning}) are common when discussing AI \cite{ye2024artificial}. Metaphorical language plays an important role in understanding complex systems \cite{lakoff2008metaphors} but the use of human metaphor can convey more humanlikeness than intended \cite{epley2007seeing}.  
There are growing concerns about anthropomorphism in systems, its implications for digital literacy and how these may lead to overreliance \cite{abercrombie-etal-2023-mirages,Akbulut_Weidinger_Manzini_Gabriel_Rieser_2024}, potentially differently between social groups.

\paragraph{Anthropomorphism.} Although some metrics have been proposed to measure anthropomorphism in the models' outputs \cite[e.g.][]{cheng-etal-2024-anthroscore}, to the best of our knowledge, no such metrics exist for user prompts. Instead, we study linguistic markers typical of human-human dialogue and metaphorical language which may be indicative of the user's mental model of the system. We measure (1) the use of politeness markers, e.g., \textit{thank you}, and phatic expressions such as \textit{hi}; (2) the use of metaphorical verbs and jargon (e.g., \textit{write} vs \textit{generate}), and 3) the use of complete sentences (e.g., ``weather in Rome'' vs.\ ``What's the weather like in Rome''), which convey a naturalness not always necessary. We use keyword spotting for (1) and (2) using manually compiled lists (more details in \Cref{app:word-lists}). For (3) we use SpaCy to find whether the prompt contains a verb. The results are shown in Table \ref{tab:anthro}. We find general trends in the data: jargon is more common for upper and middle class participants, while metaphorical language and phatic expressions are more common among lower SES participants. However, we do not find these differences to be statistically significant.

\begin{table}[]
    \centering
    \resizebox{\columnwidth}{!}{
    \begin{tabular}{r| c c c c}
    \toprule
      & Jargon & Metaphor & Phatic &  Verbs \\ \midrule
     Low & 3.32 & 25.07 &  6.34  & 63.04\\
     Middle & 4.16 & 23.72 &  5.08 & 64.98\\
     High & 4.94 & 23.49  &  4.29 & 63.42\\\bottomrule
    \end{tabular}}
    \caption{Mean percentage of prompts that contain jargon, metaphors, phatic expressions, and verbs.}
    \label{tab:anthro}
\end{table}

\paragraph{Search engine questions.}

We investigate the extent to which individuals are replacing the use of search engines with LLMs. As a proxy, we find that a notable proportion of \googlelow{}\%, \googlemiddle{}\% and \googleupper{}\% of the prompts written by people with low, middle and upper socioeconomic background respectively contains at least one of the question words: ``who'', ``what'', ``when'', ``where'', ``why'', ``how''.
In this case, the trend is that people from the low and upper classes tend to make slightly more usage of the LLMs with questions like `Which country celebrates new years first?' or `What is the difference between Espresso and regular whole bean coffee?'.

\section{Discussion and Future Directions}\label{sec:discussion}

We show that the adoption rates and use of language technologies vary significantly based on the SES of the users. 
In terms of contexts where the interactions take place, we find that mid- and higher-SES participants use LLMs more commonly for work and education. These differences may be explained by matters of access, digital literacy or \textit{habitus}, but they may exacerbate existing inequalities. 
In a recent report, \citet{unesco2023ai} has brought attention to this gap, and 
\citet{capraro2024impact} posit that the AI Gap will lead to further inequality, as certain communities benefit more from the advantages of language technologies, while already marginalised communities are increasingly left behind. For example, the benefits of generative AI in the workplace are centred around middle-class jobs, as shown in our results and recently reported in a recent report by Anthropic on the economic tasks of LLMs \cite{handa2025economic}.

We also find potential for concern in terms of the robustness of current evaluation benchmarks. The  applications associated with higher SES participants (such as paraphrasing, summarizing, and mathematical problems) are generally suited for ground-truth evaluations  \cite[e.g.][]{hendrycksmeasuring}, while the tasks more often reported by lower SES uses rely more heavily on human preference evaluation. 
Our results also support the notion that there are significant linguistic differences between groups of different SES. Although in recent years there has been growing interest in human-centred evaluation \cite[e.g.][]{xiao24humancentered,blodgett-etal-2024-human,ibrahim2024beyond}, participatory design \cite[such as][]{caselli-etal-2021-guiding} and perspectivism \cite{Frenda2024},  currently no benchmarks exist to quantify how SES differences affect NLP systems or what their real-world potential impact may be. Future work should focus on benchmarking model performance in realistic scenarios that represent the full socioeconomic spectrum, aiming to create resources and systems that address and mitigate the digital divide in NLP technologies.

\section{Conclusion}
We survey the usage of language technologies among individuals with different SES. We find statistically significant differences both in the adoption of language technologies and in the specific uses people give them. In particular, we find that upper class individuals have access to a wider variety of digital devices, use AI chatbots more frequently and with the goal to improve their work through more technical tasks like coding, data analysis or writing. We collect \totprompts{} prompts from previous interactions of our participants with LLMs, where we find statistically significant differences in the length and concreteness level across SES groups. From a qualitative analysis, we find further differences in the topics and framings of the prompts, and in the user perceptions of the systems (i.e., anthropomorphism). Our work calls for the development of inclusive NLP technologies to accommodate different SES needs and habitus and mitigate the existing AI Gap.

\section*{Limitations}
Our study is limited to U.S.- and U.K.-based crowdworkers on the Prolific platform, and may not be representative of the broader population. In terms of socioeconomic status, we expect the Prolific population to be skewed towards the middle to low social class. Furthermore, given crowdworkers' familiarity with technology, they may be more likely to use language technologies than the general population. Crowdworkers may use LLMs to complete the survey itself, e.g., by generating ten example prompts rather than providing their own, despite platform policies against the use of LLMs. 
We use the Macarthur Scale for measuring SES: Subjective metrics are prone to ambiguity and bias – with most people comparing themselves to their peers and thinking of themselves as being somewhere in the middle. However, we also note that one’s perception of their SES plays an important role in behavior and attitudes and for this reason we chose to use it in this survey.

\section*{Ethical Considerations}
\label{sec:ethics}
The survey was approved by the ethics board of the IT University of Copenhagen. Crowdworkers were compensated for their time in accordance to the platform's recommendation of  £9/hour. They may withdraw from the study by contacting the researchers. Subjects were fully anonymized in compliance with GDPR, and could opt out of supplying sensitive sociodemographic information.

\section*{Acknowledgments}
We thank the MilaNLP group in Bocconi University and the NLPnorth group at ITU for feedback on earlier version of this draft, as well as the reviewers for their helpful comments. A special thanks to Mike Zhang for help with the design of Figure 1.
Elisa Bassignana is supported by a research grant (VIL59826) from VILLUM FONDEN.
Dirk Hovy was supported by the European Research Council (ERC) under the European Union’s Horizon 2020 research and innovation program (grant agreement No.\ 949944, INTEGRATOR). He is a member the Data and Marketing Insights Unit of the Bocconi Institute for Data Science and Analysis (BIDSA).

\bibliography{anthology,custom}

\appendix

\clearpage

\section{Survey Interface}
\label{app:survey}

In \Cref{fig:survey-interface} we report the interface of our survey, including all the questions.

\section{Demographic Statistics}
\label{app:demographics}

\begin{figure}
    \centering
    \includegraphics[width=\columnwidth]{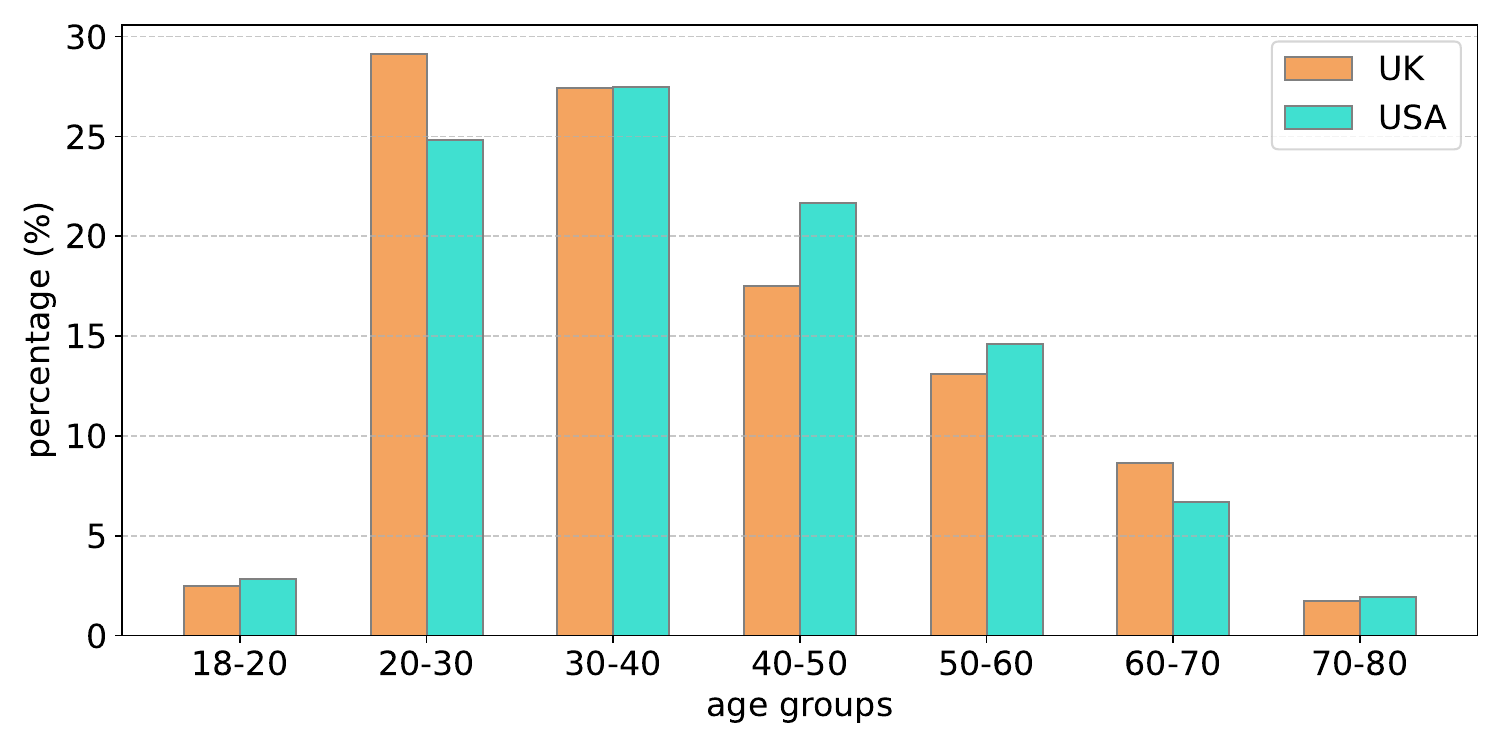}
    \caption{Distribution of participants according to the age group and divided by country of residence.}
    \label{fig:distribution-age}
\end{figure}

\begin{figure}
    \centering
    \includegraphics[width=0.9\columnwidth]{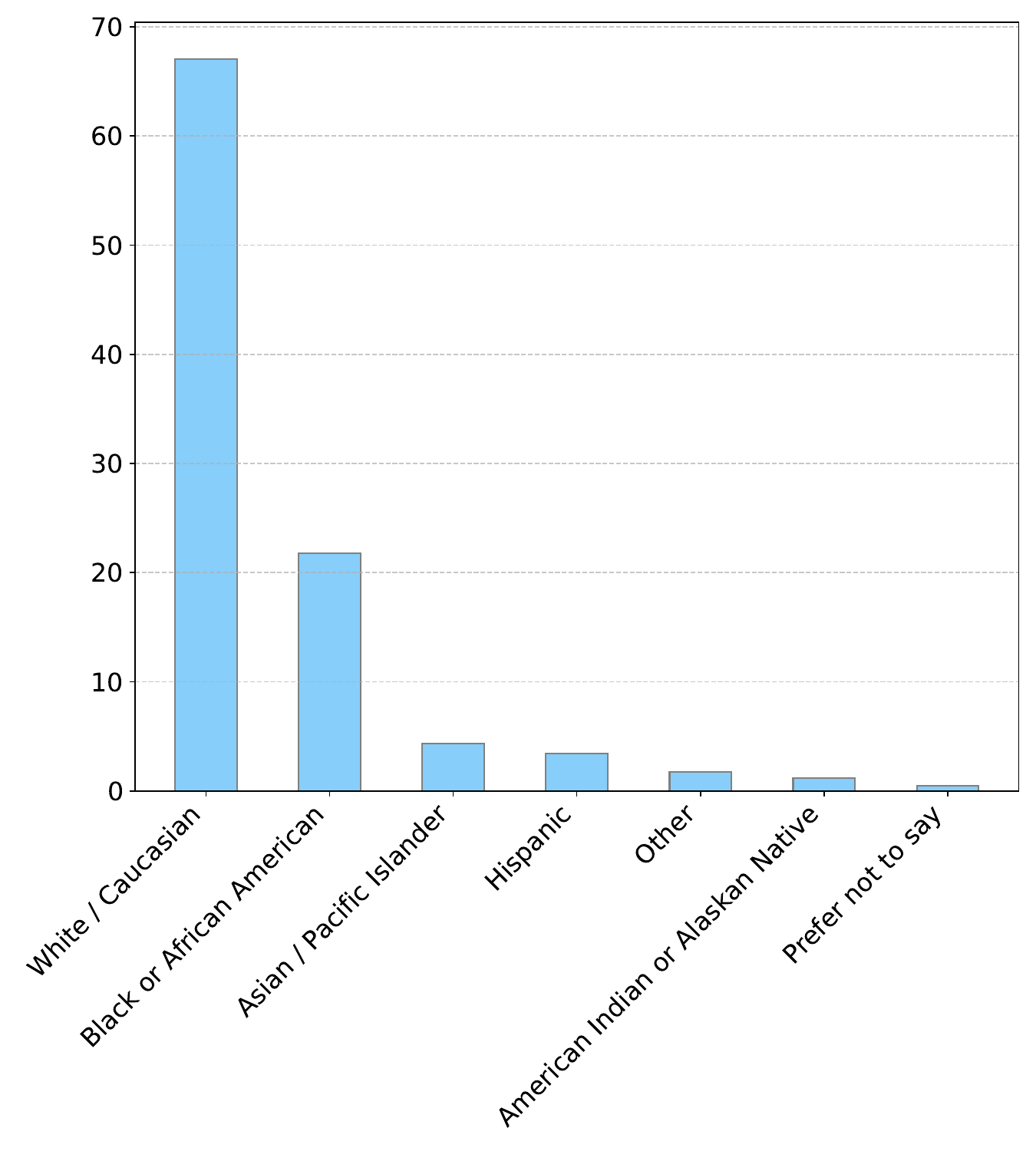}
    \caption{Distribution of participants' ethnicity.}
    \label{fig:ethnicity}
\end{figure}

\begin{figure}
    \centering
    \includegraphics[width=0.9\columnwidth]{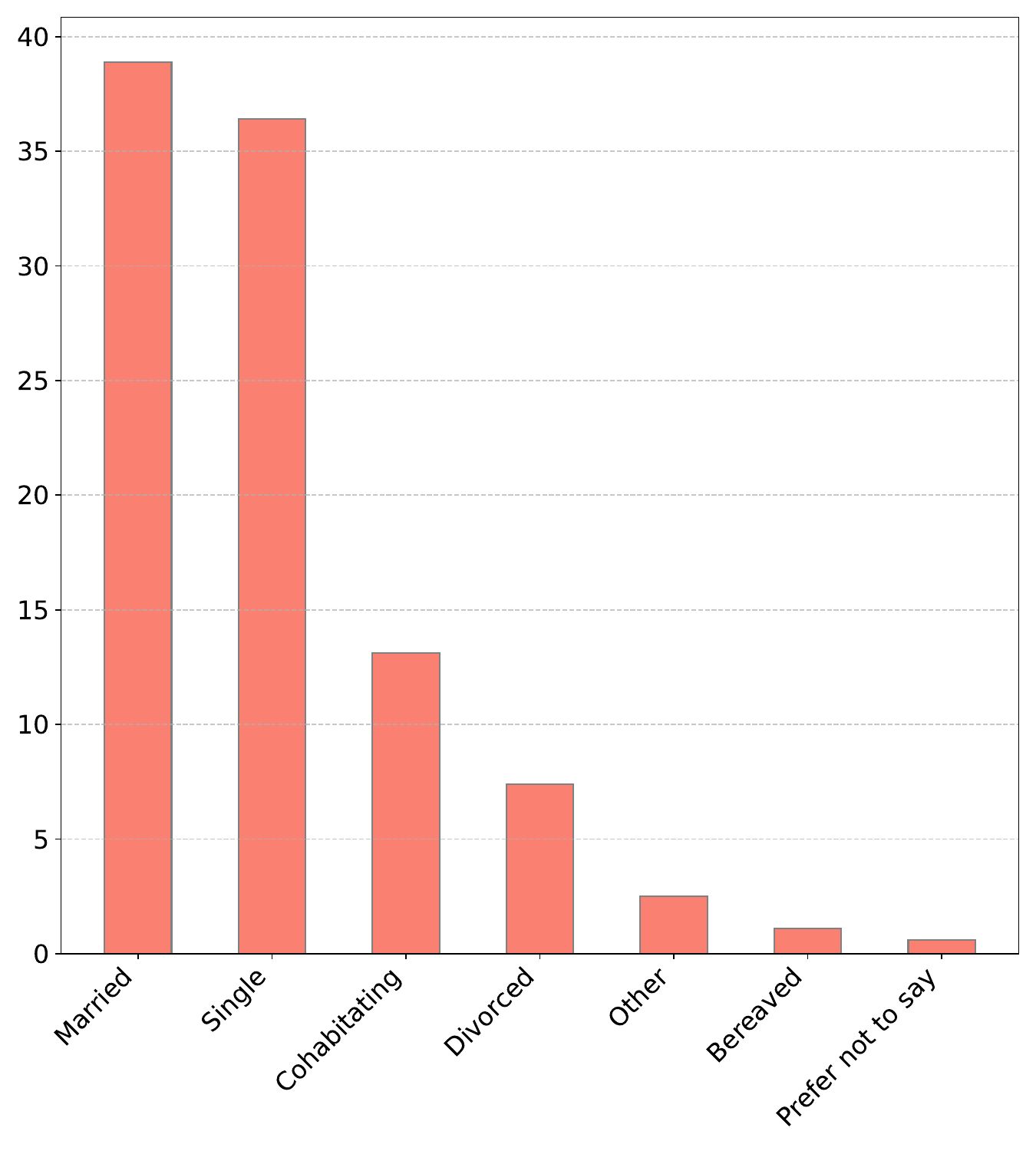}
    \caption{Distribution of participants' marital status.}
    \label{fig:marital}
\end{figure}

\begin{figure}
    \centering
    \includegraphics[width=0.9\columnwidth]{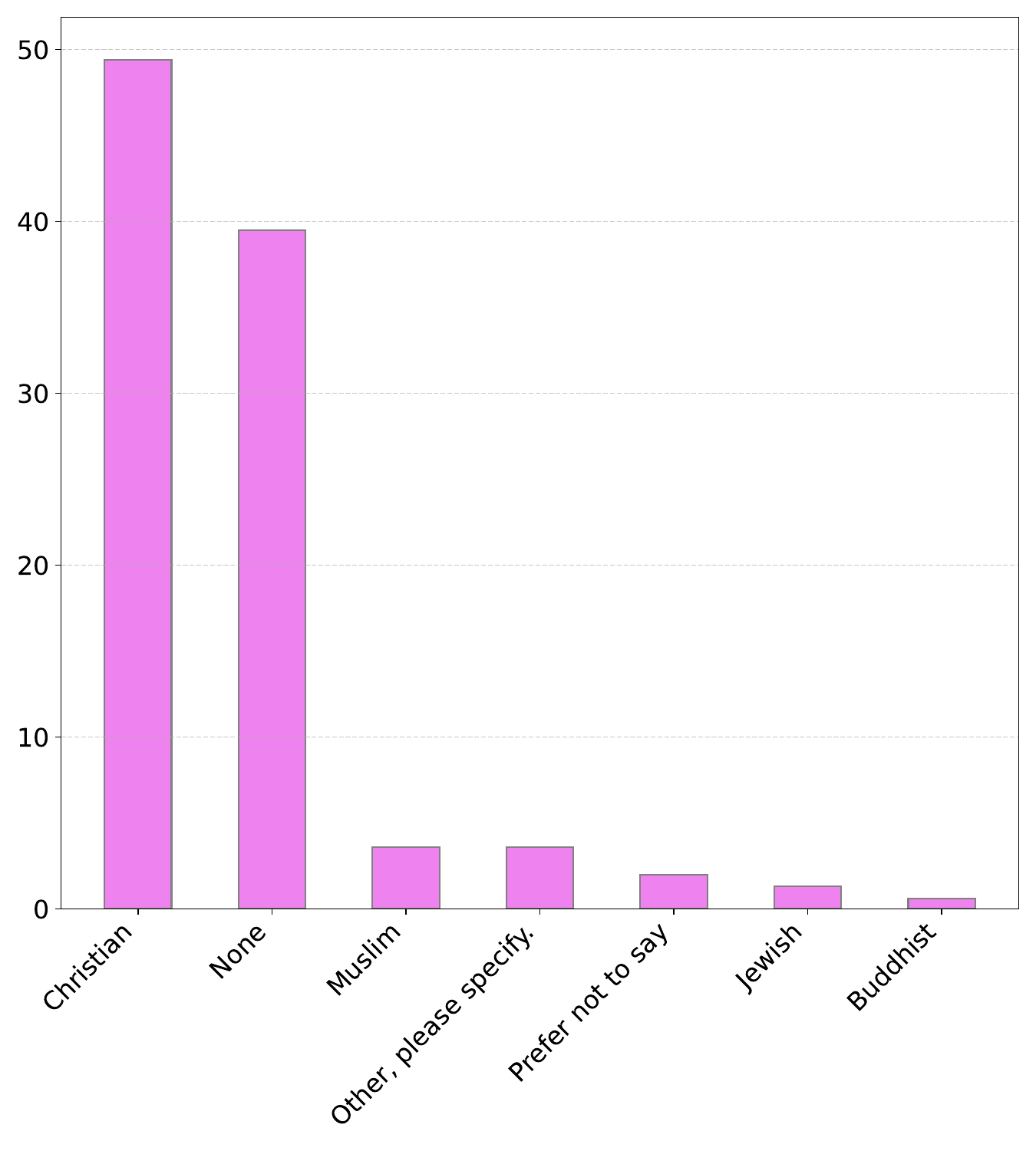}
    \caption{Distribution of participants' religion.}
    \label{fig:religion}
\end{figure}

\begin{figure}
    \centering
    \includegraphics[width=\columnwidth]{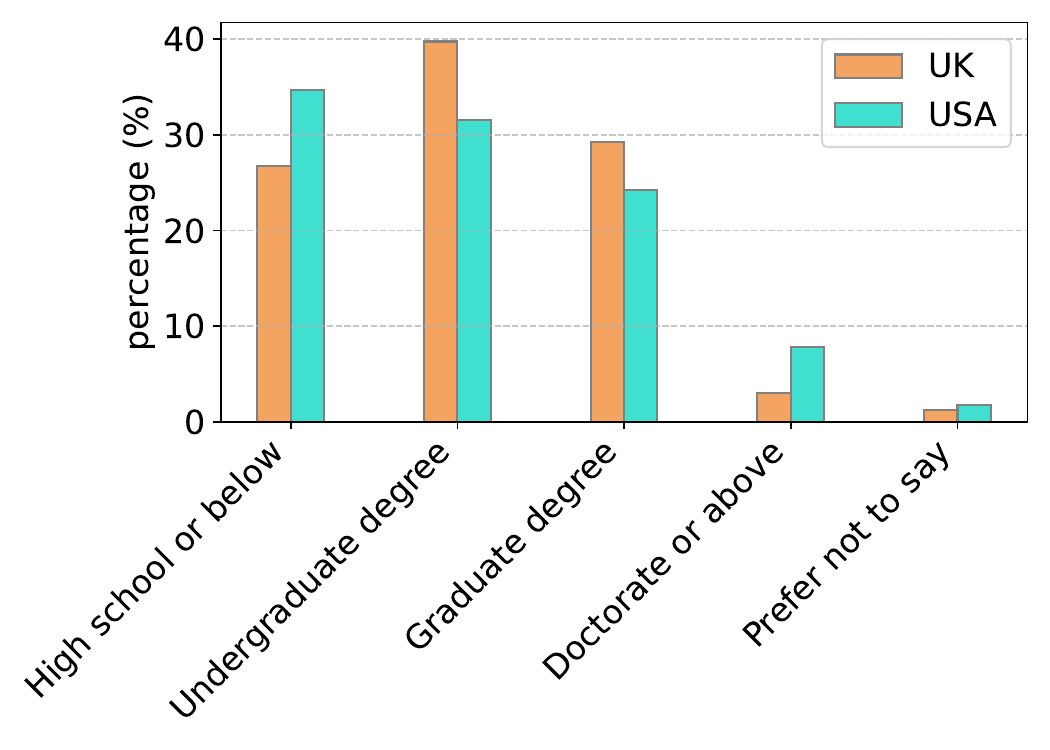}
    \caption{Distribution of participants according to the level of education and divided by country of residence.}
    \label{fig:education}
\end{figure}

\begin{figure}
    \centering
    \includegraphics[width=\columnwidth]{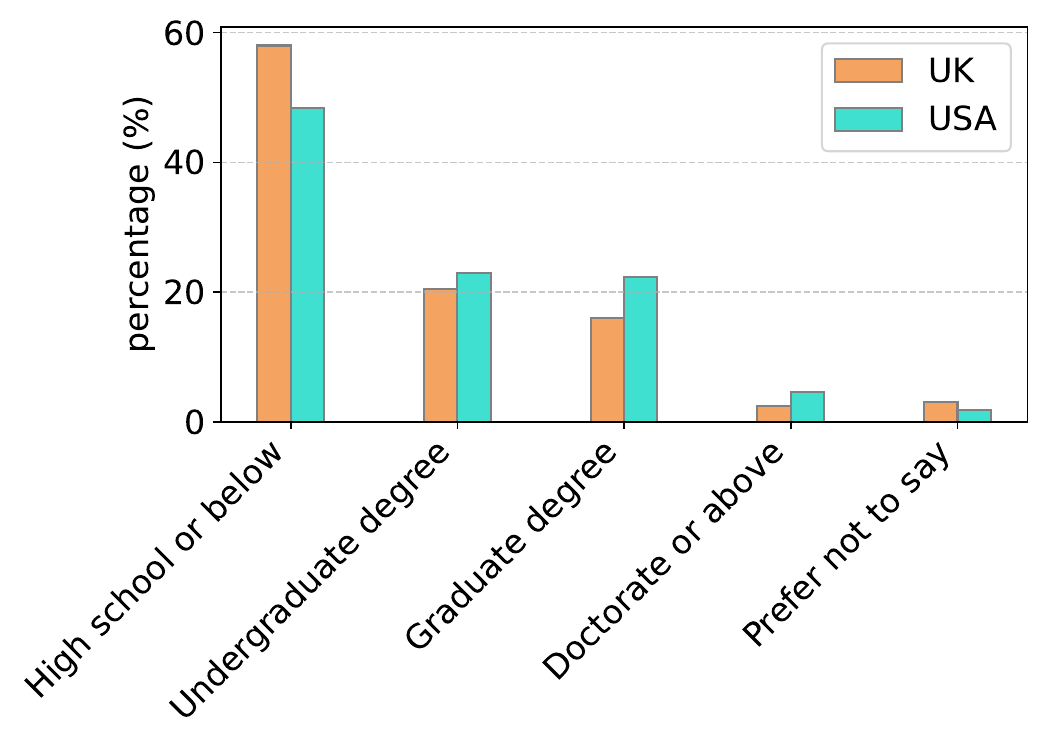}
    \caption{Distribution of participants according to the mother's level of education and divided by country of residence.}
    \label{fig:mum_education}
\end{figure}

\begin{figure}
    \centering
    \includegraphics[width=\columnwidth]{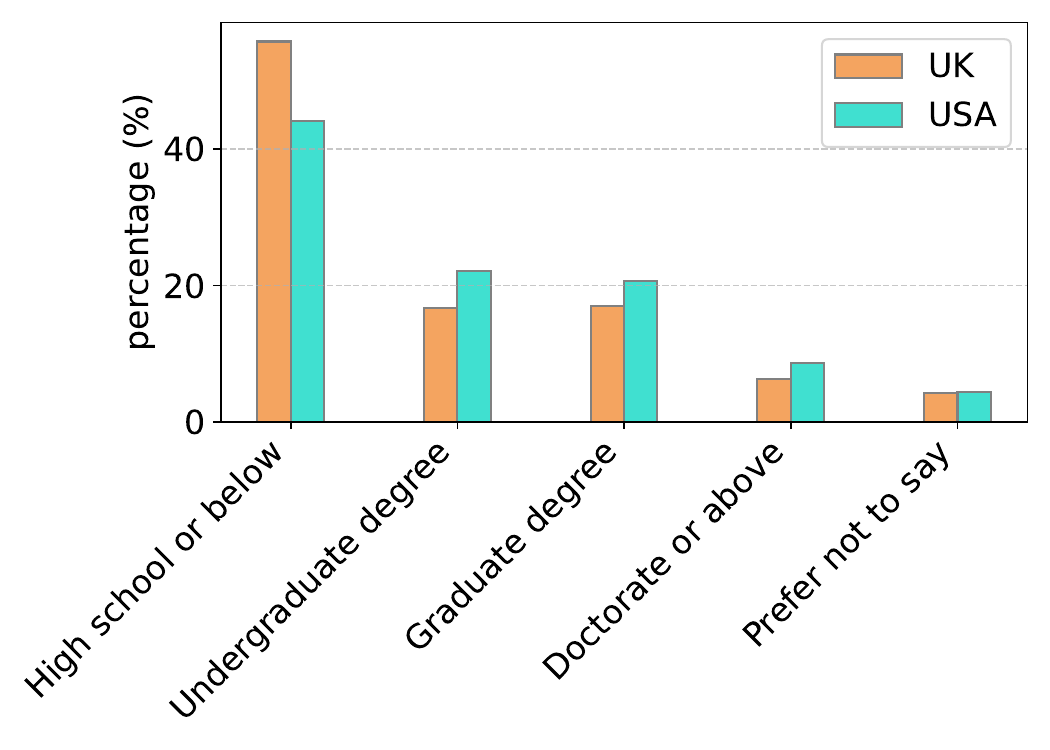}
    \caption{Distribution of participants according to the father's level of education and divided by country of residence.}
    \label{fig:dad_education}
\end{figure}

\begin{figure}
    \centering
    \includegraphics[width=\columnwidth]{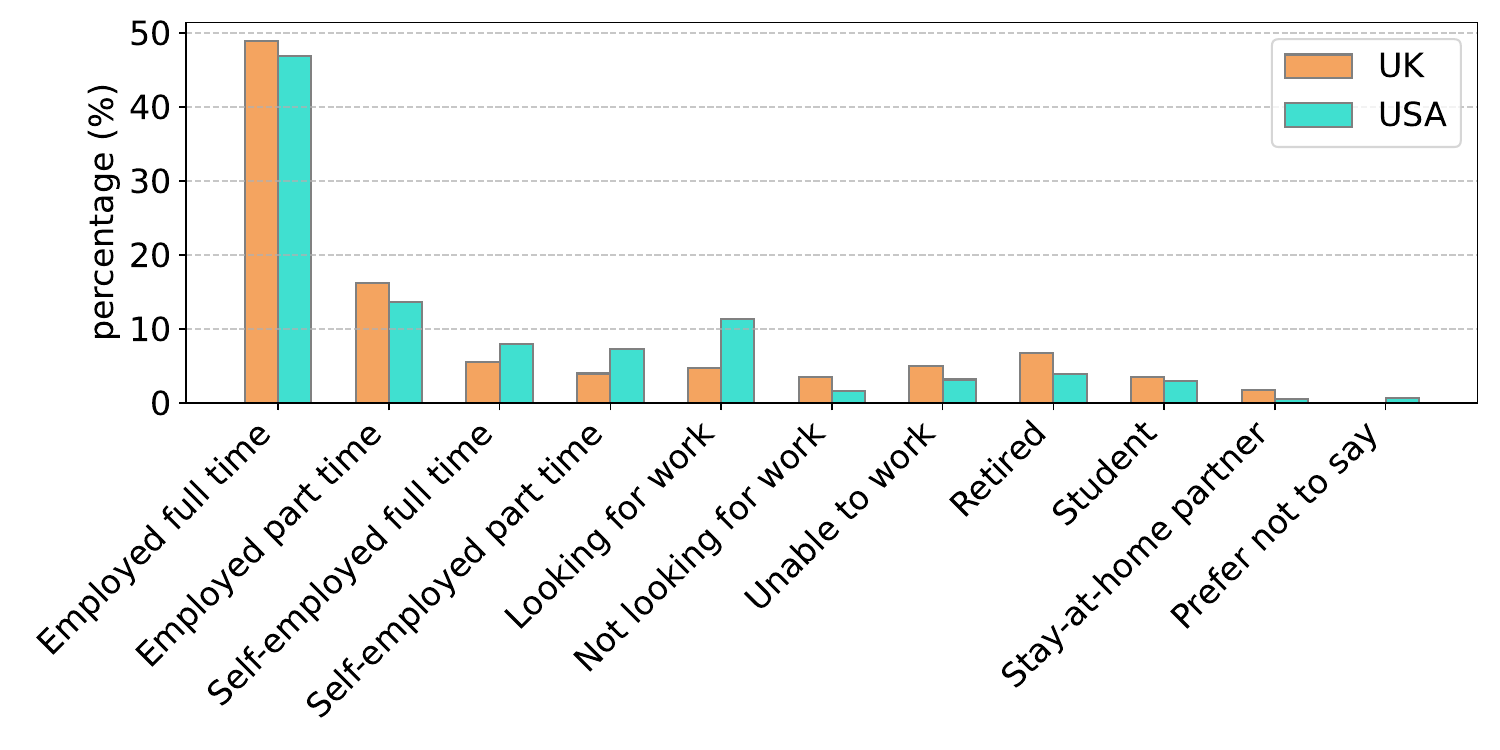}
    \caption{Distribution of participants according to the employment status and divided by country of residence.}
    \label{fig:employment}
\end{figure}

\begin{figure}
    \centering
    \includegraphics[width=\columnwidth]{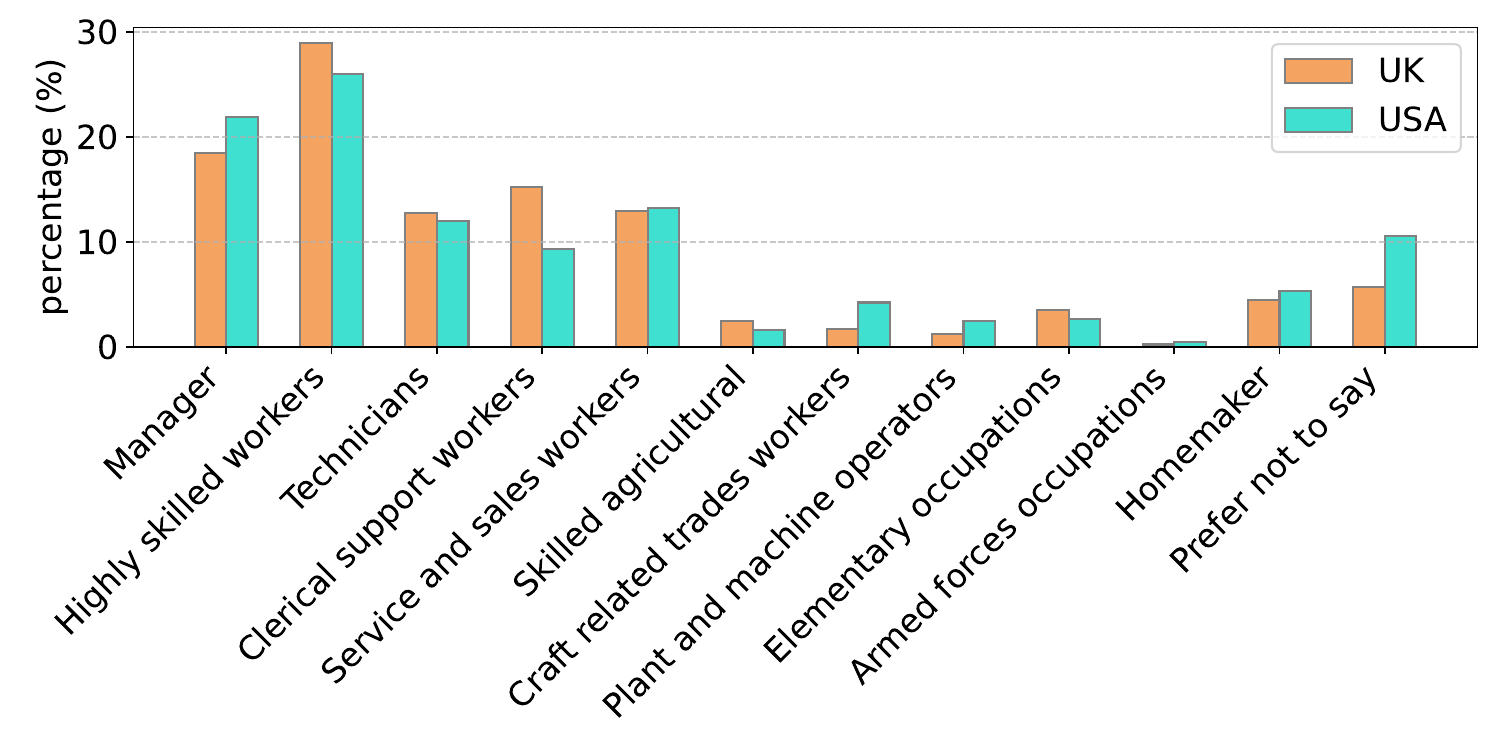}
    \caption{Distribution of participants according to the occupation and divided by country of residence.}
    \label{fig:occupation}
\end{figure}

\begin{figure}
    \centering
    \includegraphics[width=\columnwidth]{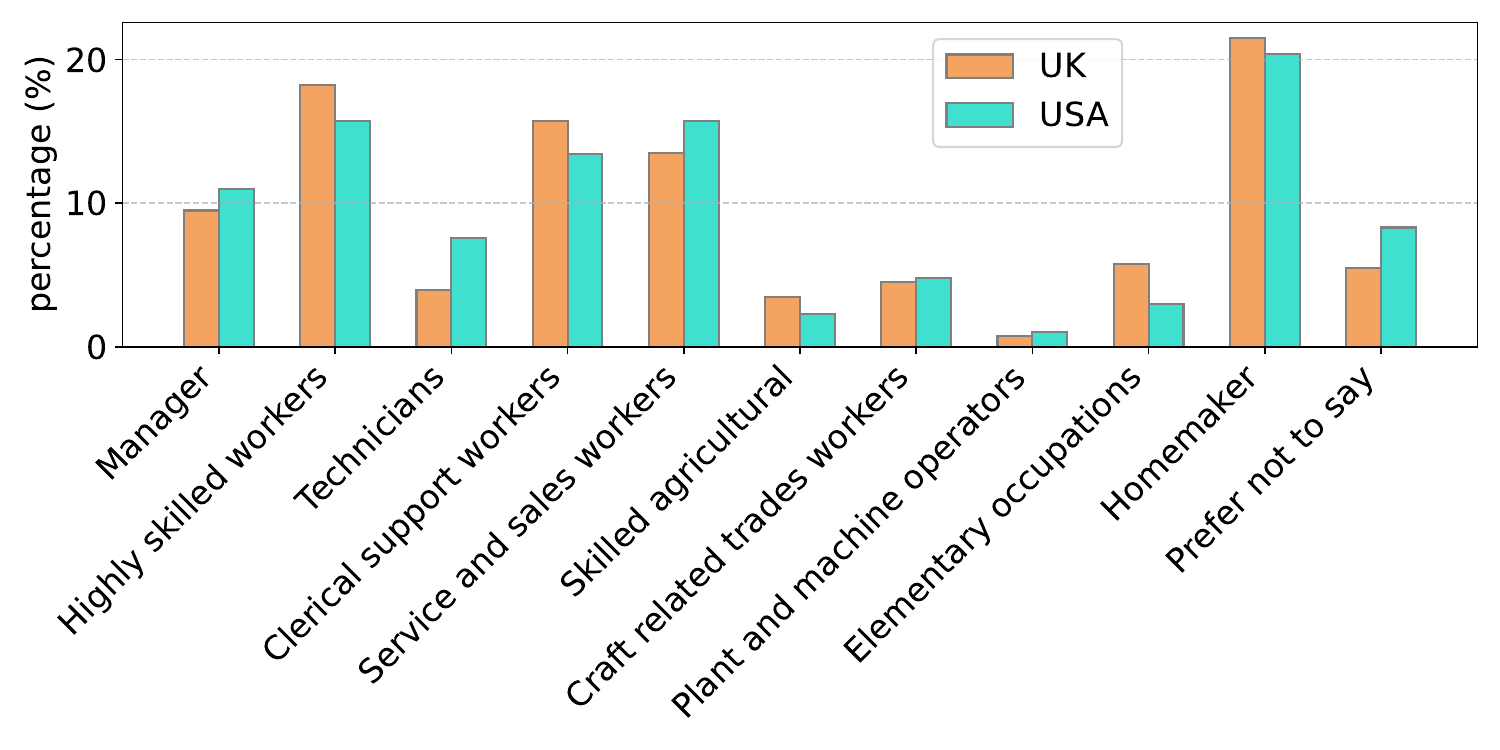}
    \caption{Distribution of participants according to the mother's occupation and divided by country of residence.}
    \label{fig:mother_occupation}
\end{figure}

\begin{figure}
    \centering
    \includegraphics[width=\columnwidth]{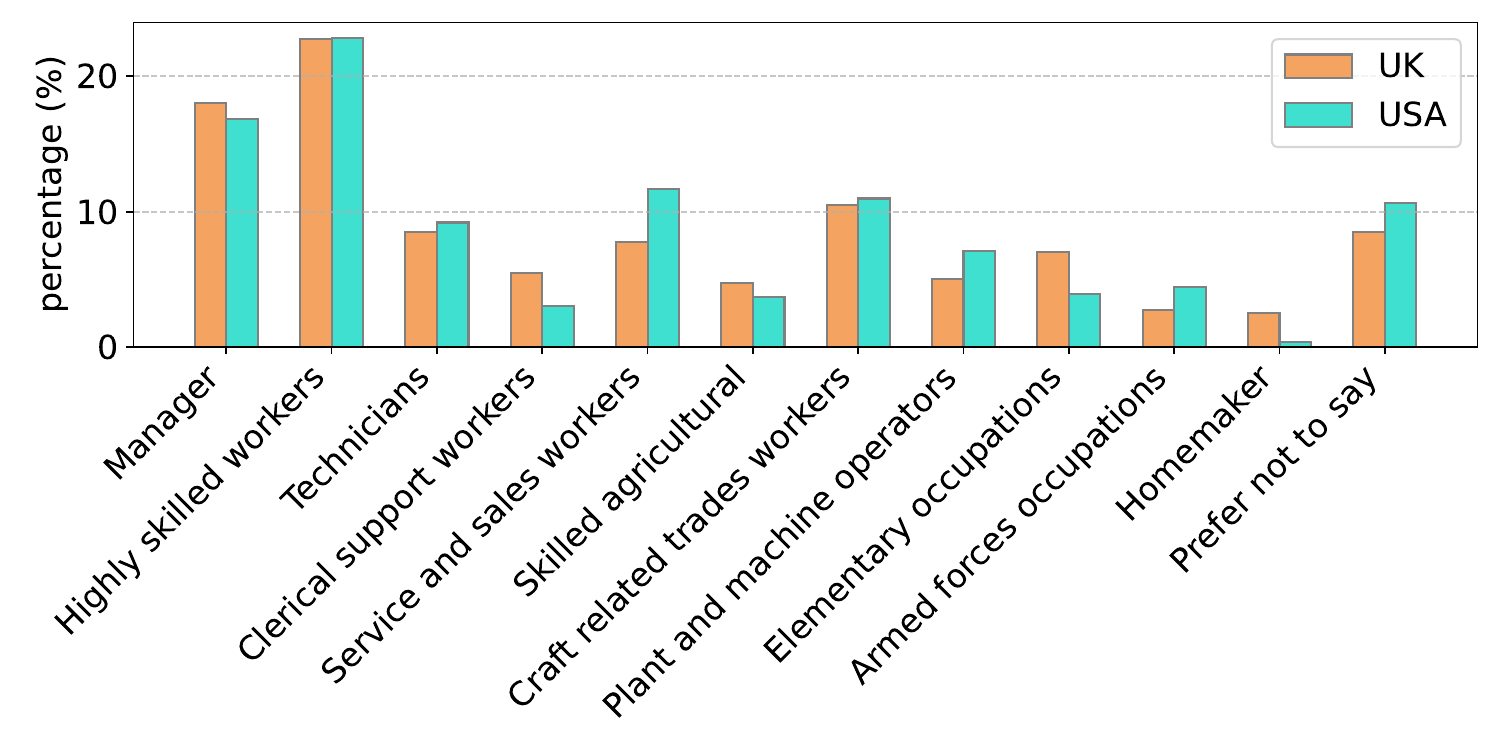}
    \caption{Distribution of participants according to the father's occupation and divided by country of residence.}
    \label{fig:father_occupation}
\end{figure}

\begin{figure}
    \centering
    \includegraphics[width=0.8\columnwidth]{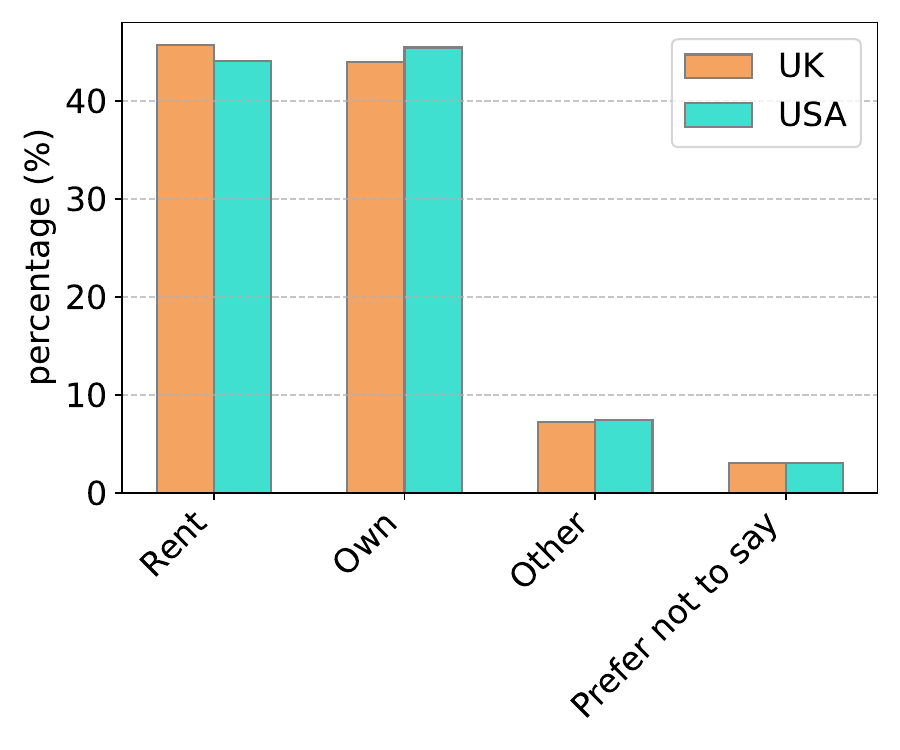}
    \caption{Distribution of participants according to the housing situation and divided by country of residence.}
    \label{fig:house}
\end{figure}

\begin{figure}
    \centering
    \includegraphics[width=\columnwidth]{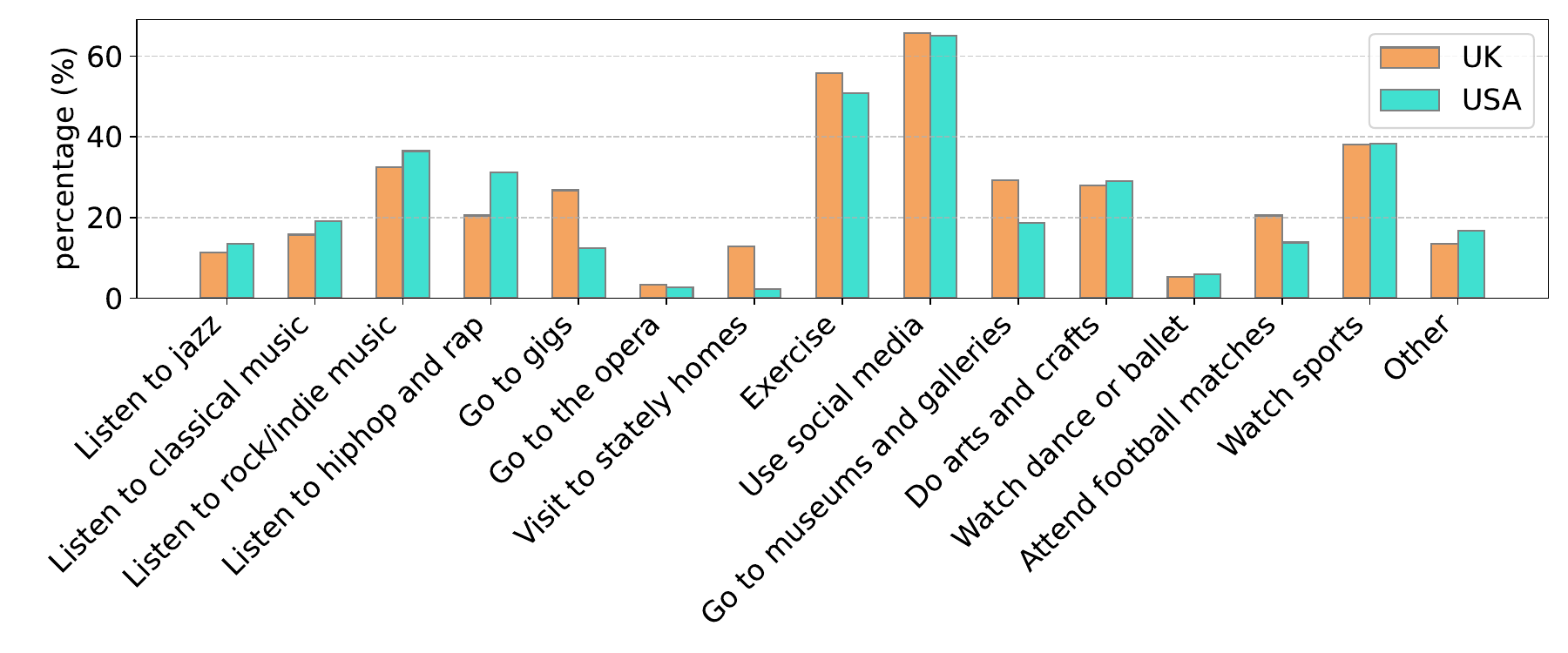}
    \caption{Distribution of participants according to the hobbies and divided by country of residence.}
    \label{fig:hobbies}
\end{figure}

We report below all the information regarding the demographics and the socioeconomic factors of our participants. 
The nationality is split as 55.3\% US, 37.8\% UK, 6.9\% `Other' including double nationality cases (mostly Canada and Nigeria). Of the total of our \totparticipants{} participants, 52.5\% identify as men, 45.6\% women, 1.6\% non-binary and 0.3\% prefer not to say.
We plot the other statistics, i.e., age (\Cref{fig:distribution-age}), ethnicity (\Cref{fig:ethnicity}), marital status (\Cref{fig:marital}), religion (\Cref{fig:religion}), level of education (\Cref{fig:education}), mother's level of education (\Cref{fig:mum_education}), father's level of education (\Cref{fig:dad_education}), employment status (\Cref{fig:employment}), occupation according to the European Skills, Competences, Qualifications and Occupations taxonomy (ESCO;~\citealp{esco}) (\Cref{fig:occupation}), mother's occupation (\Cref{fig:mother_occupation}), father's occupation (\Cref{fig:father_occupation}), housing situation (\Cref{fig:house}) and hobbies, following the list provided by Great British Class Survey~\cite{gbcs} (\Cref{fig:hobbies}).

\section{Language Technologies}
\label{app:language-technologies}

\begin{figure}
    \centering
    \includegraphics[width=\columnwidth]{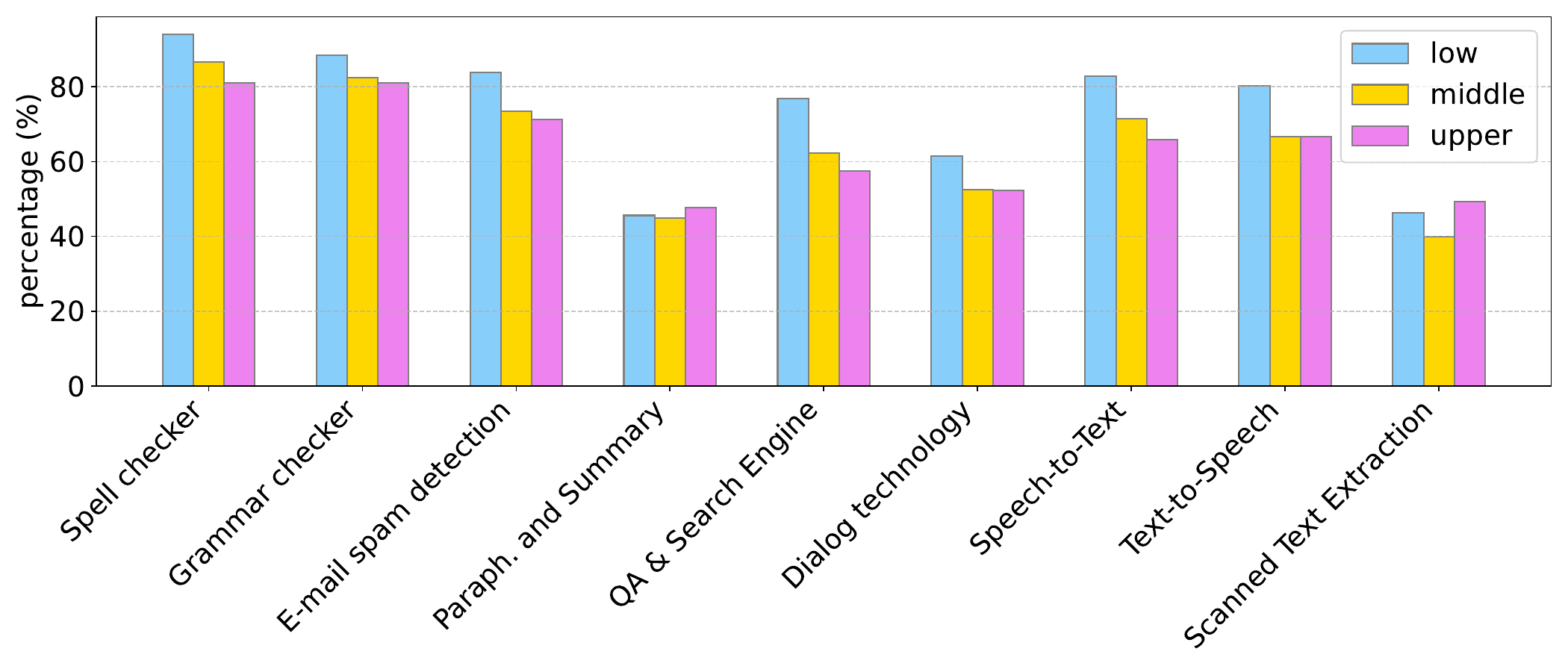}
    \caption{Type of language technologies known by individual from low, middle, upper classes.}
    \label{fig:know_by_ses}
\end{figure}

We report in \Cref{fig:know_by_ses} the statistics related to the question `Which of the following language technologies have you heard about?'. The distributions across the low, middle and upper classes are statistically significant, as indicated by a chi-square test of independence, $\chi^2$ (df = 24, N = 6449) = 166.44, \textit{p} < 0.001. 

\begin{figure}
    \centering
    \includegraphics[width=\columnwidth]{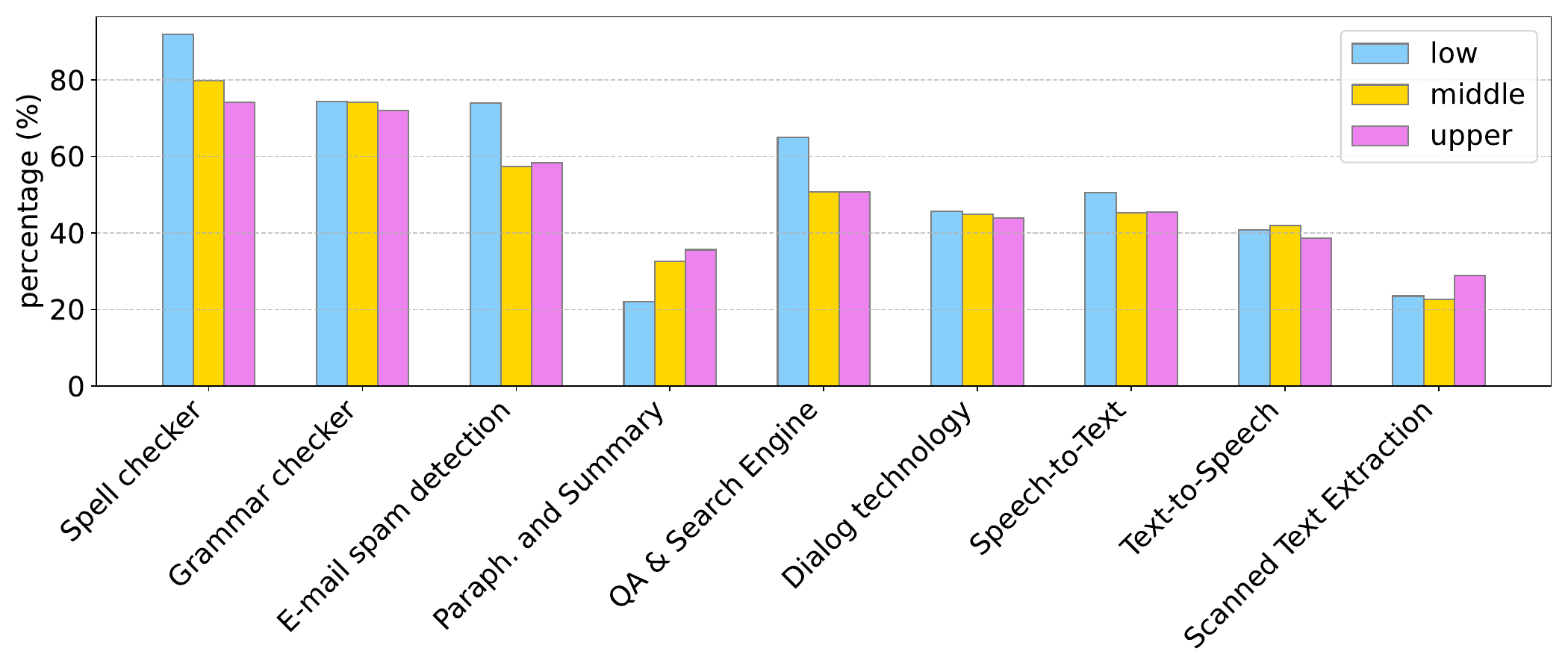}
    \caption{Type of language technologies used by individual from low, middle, upper classes.}
    \label{fig:use_by_ses}
\end{figure}

In \Cref{fig:use_by_ses} we report the statistics relative to the question `Which of the following language technologies have you used?'. The distributions across the low, middle and upper classes are statistically significant, as indicated by a chi-square test of independence, $\chi^2$ (df = 24, N = 4889) = 166.34, \textit{p} < 0.001.

\begin{figure}
    \centering
    \includegraphics[width=\columnwidth]{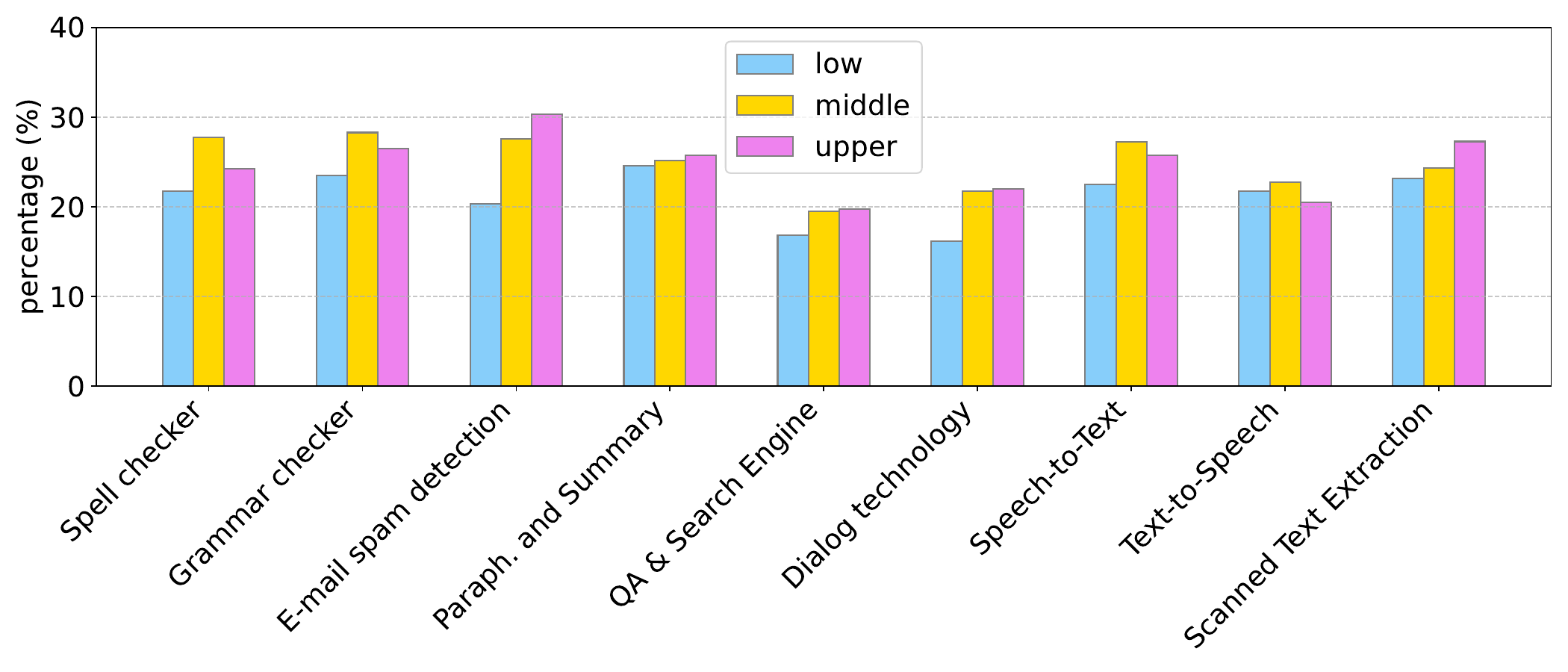}
    \caption{Type of language technologies that individual from low, middle, upper classes would like to use, but do not perform well enough.}
    \label{fig:would_by_ses}
\end{figure}

\Cref{fig:would_by_ses} reports the statistics relative to the question `Below is a list of some common language technologies. Please check every one that you would find useful, but do not use because of scarce performance'. The distributions across the low, middle and upper classes are statistically significant, as indicated by a chi-square test of independence, $\chi^2$ (df = 24, N = 2490) = 115.65, \textit{p} < 0.001.

\begin{figure}
    \centering
    \includegraphics[width=\columnwidth]{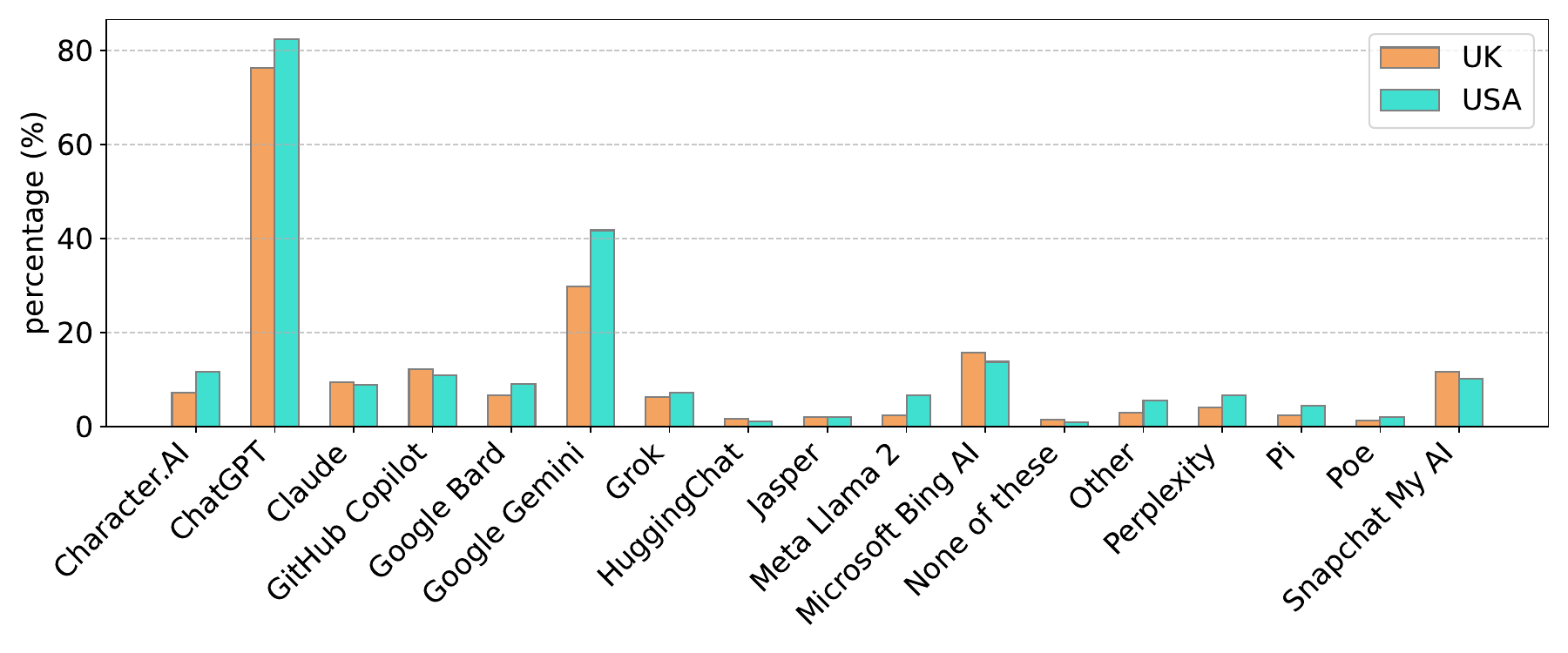}
    \caption{Type of LLMs used by individual divided by country of residence.}
    \label{fig:llm_use}
\end{figure}

Last, in \Cref{fig:llm_use} we focus on the usage of LLMs and report the statistics relative to the question `If you use them, which of these AI chatbots do you use? If you have never used them, leave blank.'.

\section{Wordify}
\label{app:wordify}

We use a variant of the Stability Selection algorithm~\cite{stability} as implemented by \citet{wordify} to extract the most indicative n-grams for each social class (low, middle, upper). We do not find significant differences across SES groups. We report the results below.

\paragraph{Positive Indicators.}

Upper: kindly (0.414), british (0.398), employee (0.38), draft (0.364), sell (0.356), team (0.348), email (0.344), plan (0.344), donald (0.332), usa (0.332), what is (0.32), column (0.316), code (0.308), task (0.306), the monthly (0.306), manager (0.302), name of (0.302), or (0.3).

Middle: datum (0.394), add (0.376), why (0.362), fun (0.36), my (0.36), analysis (0.352), car (0.348), question (0.344), you know (0.342), hello (0.334), love (0.334), business (0.33), air (0.324), build (0.324), climate (0.322), website (0.322), cv (0.312), if you (0.312), which (0.312), customer (0.31), js (0.31), made (0.31), earth (0.308), order (0.306), holiday (0.304), meaning (0.304), quality (0.304), service (0.304), summarise (0.304), tiktok (0.3), true (0.3).

Low: how to (0.42), with (0.414), day (0.406), video (0.4), was (0.396), where (0.394), need (0.39), be (0.384), make (0.384), it (0.38), summarize (0.374), do (0.358), off (0.352), chicken (0.35), tell me about (0.346), description (0.344), total (0.344), resume (0.334), to cook (0.334), idea for (0.332), art (0.33), name (0.326), something (0.326), by (0.322), is (0.322), image (0.32), is the good (0.32), is there (0.32), someone (0.316), fill (0.314), tall (0.31), animal (0.306), country (0.306), how (0.304), ask (0.302), did (0.302), would (0.302), cheap (0.3), those (0.3), trade (0.3).

\paragraph{Negative Indicators.}

Upper: my (0.422), make (0.414), can (0.388), an (0.38), create (0.374), me (0.356), of (0.356), which (0.342), people (0.334), about (0.322), question (0.312), where (0.3).

Middle: used (0.408), summarize (0.406), ask (0.392), with (0.392), is there (0.38), is (0.376), trump (0.376), video (0.374), day (0.364), it (0.36), will (0.358), her (0.356), what are (0.356), to write (0.354), those (0.35), sell (0.34), hour (0.336), new (0.336), not (0.332), generate me (0.326), horror (0.326), idea for (0.322), how (0.32), well (0.32), creation (0.316), legal (0.316), country (0.312), do not (0.31), employee (0.31), fact (0.31), would (0.31), chicken (0.308), can use (0.306), or (0.306), you (0.306), how to make (0.304), up (0.304), play (0.302), trade (0.302).

Low: company (0.422), analysis (0.396), improve (0.386), you help (0.38), function (0.372), how does (0.372), datum (0.37), love (0.352), project (0.346), climate (0.338), if you (0.328), what is (0.324), business (0.322), are the (0.32), time (0.316), such (0.308), and (0.3), change (0.3).

\section{Anthropomorphism} 
\label{app:word-lists}

Below we report the list of words used for the analysis of user perceptions with respect to anthropomorphism.

Phatic expression: thank, thanks, please, hi, hello.

Metaphorical verbs: ask, assess, care, choose, create, decide, describe, discover, empathize, engage, enjoy, evaluate, explain, express, feel, find, hear, imagine, improve, invent, judge, know, learn, listen, look, observe, plan, predict, prioritize, rate, react, reason, recommend, remember, respond, search, see, solve, speak, suggest, think, translate, understand, watch, write

Jargon: activate, compute, detect, evaluate, forecast, generate, ingest, infer, input, map, monitor, optimize, output, parse, prioritize, process, query, rank, rationalize, register, render, resolve, respond, retrieve, score, select, simulate, store, synthesize, track.

\clearpage

\begin{figure}[t]
    \centering
    \begin{subfigure}[t]{0.5\textwidth}
        \centering
        \includegraphics[width=\linewidth]{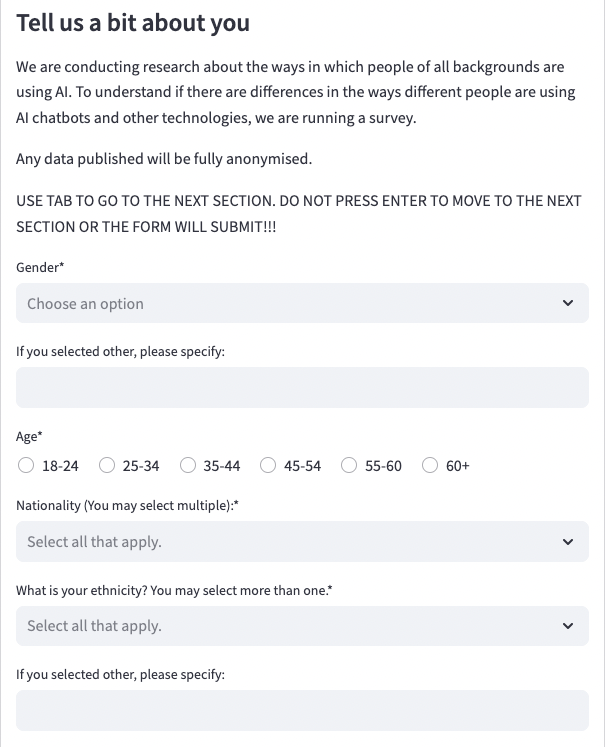}
    \end{subfigure}
    \hfill
    \begin{subfigure}[t]{0.5\textwidth}
        \centering
        \includegraphics[width=\linewidth]{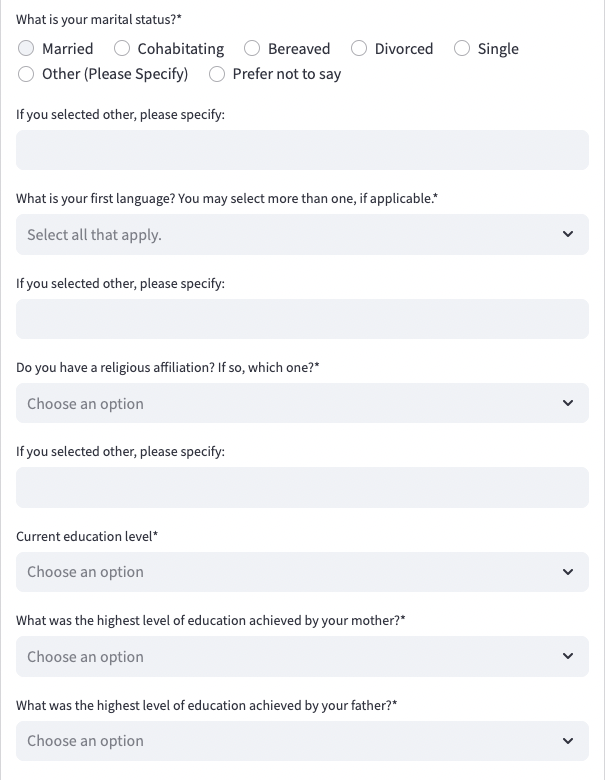}
    \end{subfigure}
\end{figure}

\begin{figure}[ht]
    \centering
    \begin{subfigure}{0.5\textwidth}
        \centering
        \includegraphics[width=\linewidth]{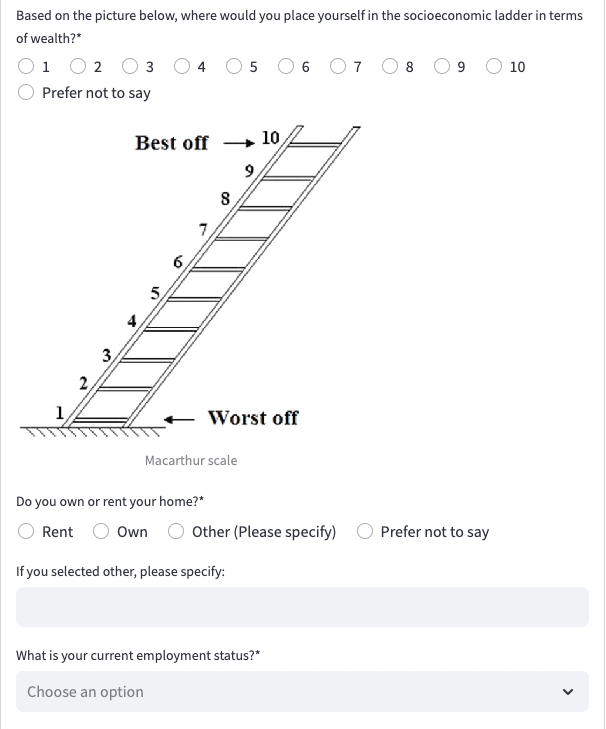}
    \end{subfigure}
    \hfill
    \begin{subfigure}{0.5\textwidth}
        \centering
        \includegraphics[width=\linewidth]{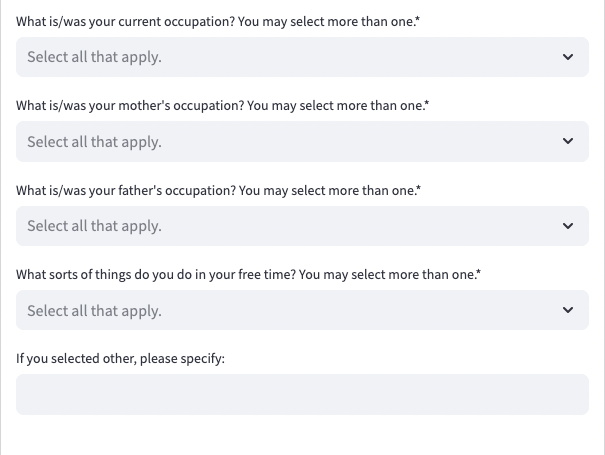}
    \end{subfigure}
    \caption*{[continue in the next page]}
\end{figure}

\begin{figure}
    \centering
    \begin{subfigure}{0.5\textwidth}
        \centering
        \includegraphics[width=\linewidth]{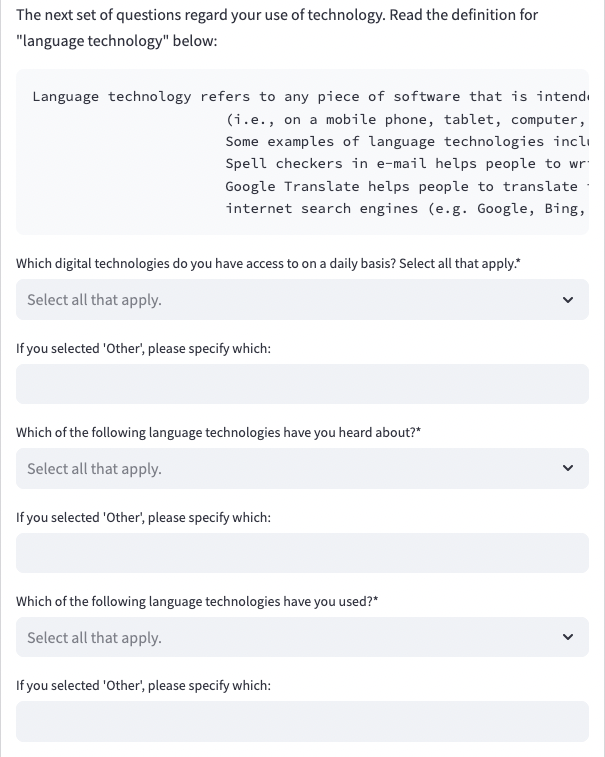}
    \end{subfigure}
    \hfill
    \begin{subfigure}{0.5\textwidth}
        \centering
        \includegraphics[width=\linewidth]{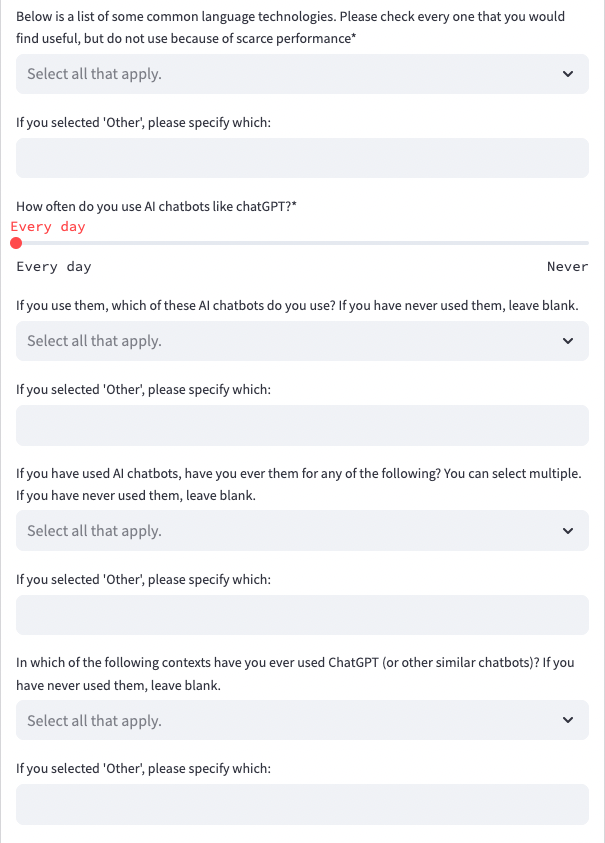}
    \end{subfigure}
\end{figure}

\begin{figure}
    \centering
    \begin{subfigure}{0.5\textwidth}
        \centering
        \includegraphics[width=\linewidth]{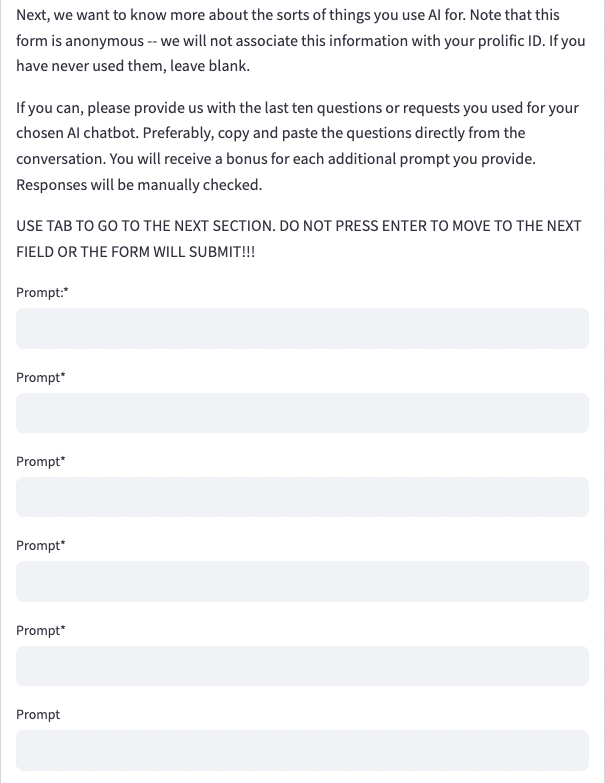}
    \end{subfigure}
    \hfill
    \begin{subfigure}{0.5\textwidth}
        \centering
        \includegraphics[width=\linewidth]{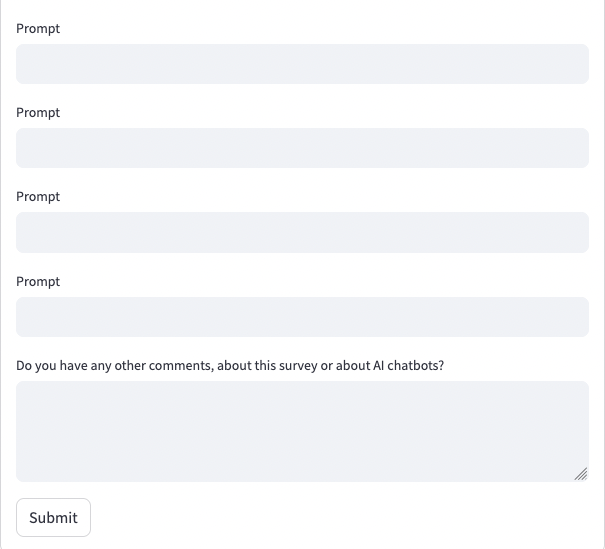}
    \end{subfigure}
    \caption{Complete interface of the survey.}
    \label{fig:survey-interface}
\end{figure}

\end{document}